\documentclass[conference]{IEEEtran}[10pt]
\IEEEoverridecommandlockouts
\usepackage{cite}
\usepackage{amsmath,amssymb,amsfonts}
\usepackage{algorithmic}
\usepackage{graphicx}
\usepackage{subcaption}
\usepackage{textcomp}
\usepackage{xcolor}
\usepackage{url}
\usepackage{color}
\usepackage{caption} %
\usepackage{multirow}
\usepackage{adjustbox}
\usepackage[normalem]{ulem}
\usepackage{flushend} 
\usepackage{titlesec}

\titlespacing*{\section}
  {0pt}  
  {6pt} 
  {3pt}  

\titlespacing*{\subsection}
  {0pt}  
  {5pt}  
  {3pt}  

\renewcommand{\thesubsubsection}{\arabic{subsubsection})}

\titleformat{\subsubsection}[runin]
  {\normalfont\normalsize\itshape} 
  {\thesubsubsection}{0.5em}{}[: \space] 

\titlespacing*{\subsubsection}
  {0pt} 
  {0pt} 
  {0pt} 
  
\def\BibTeX{{\rm B\kern-.05em{\sc i\kern-.025em b}\kern-.08em
    T\kern-.1667em\lower.7ex\hbox{E}\kern-.125emX}}

\newcommand{\commentss}[1]{}
\begin{document}

\title{I Can't Believe It's Not Real: CV-MuSeNet: Complex-Valued Multi-Signal Segmentation}
\author{\IEEEauthorblockN{Sangwon Shin and Mehmet C. Vuran}
\IEEEauthorblockA{\textit{Cyber-Physical Networking Lab, School of Computing} \\
University of Nebraska-Lincoln, Lincoln, Nebraska, USA \\
\{sshin11, mcv\}@unl.edu}\thanks{This work is supported by the Department of Navy, Office of Naval Research, NSWC N00174-23-1-0007 grants. This work relates to the Department of Navy award N00174-23-1-0007 issued by the Office of Naval Research. Any opinions, findings, and conclusions or recommendations expressed in this material are those of the authors and do not necessarily reflect the views of the Office of Naval Research.}
}

\maketitle
\begin{abstract}
The increasing congestion of the radio frequency spectrum presents challenges for efficient spectrum utilization. Cognitive radio systems enable dynamic spectrum access with the aid of recent innovations in neural networks. However, traditional real-valued neural networks (RVNNs) face difficulties in low signal-to-noise ratio (SNR) environments, as they were not specifically developed to capture essential wireless signal properties such as phase and amplitude. This work presents $\mathbb{C}$MuSeNet, a complex-valued multi-signal segmentation network for wideband spectrum sensing, to address these limitations. 

Extensive hyperparameter analysis shows that a naive conversion of existing RVNNs into their complex-valued counterparts is ineffective. Built on complex-valued neural networks (CVNNs) with a residual architecture, $\mathbb{C}$MuSeNet introduces a complex-valued Fourier spectrum focal loss ($\mathbb{C}$FL) and a complex plane intersection over union ($\mathbb{C}$IoU) similarity metric to enhance training performance. Extensive evaluations on synthetic, indoor over-the-air, and real-world datasets show that $\mathbb{C}$MuSeNet achieves an average accuracy of 98.98\%-99.90\%, improving by up to 9.2 percentage points over its real-valued counterpart and consistently outperforms state of the art. Strikingly, $\mathbb{C}$MuSeNet achieves the accuracy level of its RVNN counterpart in just two epochs, compared to the 27 epochs required for RVNN, while reducing training time by up to a 92.2\% over the state of the art. The results highlight the effectiveness of complex-valued architectures in improving weak signal detection and training efficiency for spectrum sensing in challenging low-SNR environments. The dataset is available at: \url{https://dx.doi.org/10.21227/hcc1-6p22}
\end{abstract}

\begin{IEEEkeywords}
spectrum sensing, cognitive radio, complex-valued neural networks, deep learning 
\end{IEEEkeywords}
\section{Introduction}
\label{sec:intro}

The rapid growth of wireless communication technologies has congested the radio frequency (RF) spectrum, creating critical challenges for efficient utilization. Traditional fixed spectrum allocation methods are insufficient to meet the surging demand from connected devices~\cite{Zhang20SpectrumAllocation}. Cognitive radio offers a promising solution by enabling dynamic access to underutilized frequency bands~\cite{Mitola99CogRadio, Sahai04cogradio, Ma09congnitive_signal_processing, Ian06DynamicSpecAccess}. 

\begin{figure}[t!]
    \centering
    \includegraphics[width=0.45\textwidth]{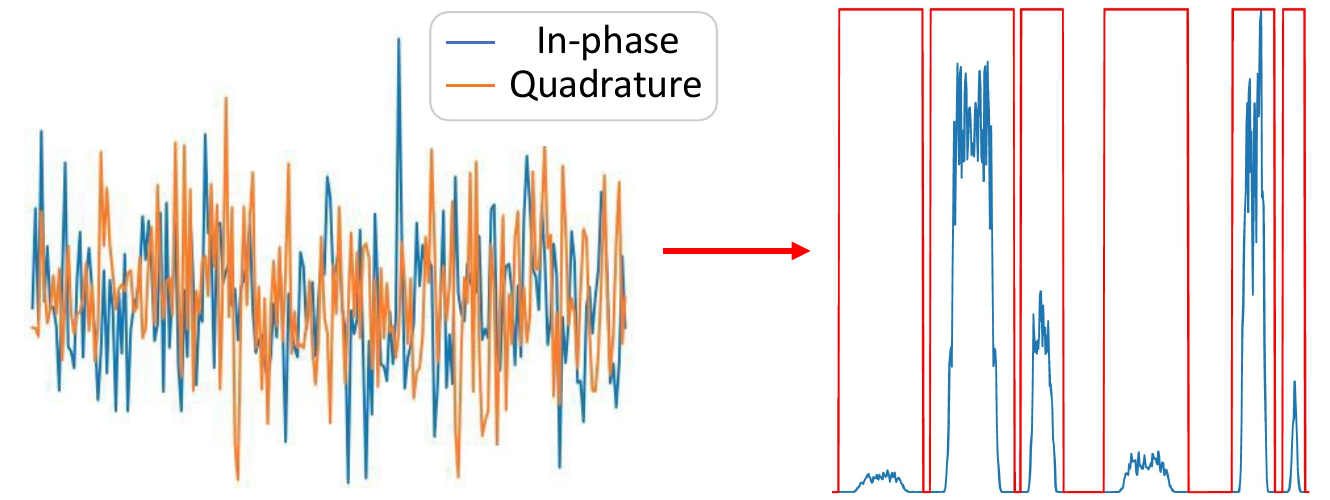}
    \caption{Multi-signal spectrum segmentation.}
    \label{fig:spec_seg_illust}\vspace{-0.2in}
\end{figure}

Detecting and segmenting signals within wideband spectrum environments, referred to as spectrum segmentation (Fig.~\ref{fig:spec_seg_illust}), is a critical challenge in cognitive radio systems~\cite{Zhao07DynamicSpecAccess, Awin19blindSpecSens}. Effective segmentation enables the identification of available spectrum bands, facilitating efficient utilization. However, real-world scenarios often involve multiple signals and low signal-to-noise ratio (SNR) conditions, which complicate this task~\cite{Oshea16_amc_simulation, Subedi24}. Traditionally, spectrum segmentation relies on real-valued neural networks (RVNNs), which process in-phase and quadrature (IQ) signals by separating them into real and imaginary components. However, RVNNs are inherently limited in capturing essential signal characteristics, such as phase and amplitude relationships, under realistic conditions. These limitations result in reduced accuracy, particularly in low-SNR environments. 

Recent advances in deep learning show that complex-valued neural networks (CVNNs) effectively learn functions in the complex domain ($f: \mathbb{C} \to \mathbb{C}$)~\cite{hirose2012complexnn}. CVNNs inherently capture the relations between real and imaginary components of a complex-valued number (e.g., phase), making them suitable for various applications~\cite{trabelsi2018deepnn}. CVNNs consistently outperform traditional methods in fields like magnetic resonant imaging (MRI), where preserving the complex-valued nature of the data is crucial. When applied to IQ signals, CVNNs have the potential to retain critical characteristics such as signal phase, phase shifts, and multipath effects, which are often approximated, rather than \textit{learned} in traditional real-valued approaches~\cite{hirose2012complexnn}. Although in their infancy, CVNNs have recently been successfully utilized in RF processing and communications, demonstrating their versatility~\cite{hirose2012complexnn, Zhao21DeepWaveform, Xu22AnalysisCVNN_RF}. 

The potential of CVNNs for spectrum segmentation in dynamic spectrum access remains unexplored. While complex-valued architectures promise enhanced learning capabilities with wireless signals, as we show in this paper, a naive conversion of existing real-valued architectures to their complex-valued counterparts has limitations. These limitations motivate the development of tailored CVNN components, such as complex-valued loss functions and learning metrics, to fully exploit CVNNs' capabilities for segmentation tasks, enabling improvements in challenging low-SNR conditions.

In this paper, we introduce $\mathbb{C}$MuSeNet, a complex-valued neural network for multi-signal spectrum segmentation. $\mathbb{C}$MuSeNet leverages CVNNs to improve segmentation accuracy while reducing training time.
This work makes the following contributions: 
\vspace{-0.025in}
\begin{itemize}
    \item We design $\mathbb{C}$MuSeNet, a CVNN-based segmentation architecture that can directly process IQ signals and their complex-valued Fourier transforms, preserving critical signal characteristics.
    \item We introduce a complex-valued Fourier spectrum focal loss ($\mathbb{C}$FL) and a complex plane intersection over union ($\mathbb{C}$IoU) metric tailored for CVNN-based  segmentation.
    \item We implement a residual CVNN architecture and complex-valued training strategies. An extensive hyperparameter analysis allows us to fine-tune $\mathbb{C}$MuSeNet to accelerate convergence and reduce training time.
    \item We demonstrate state-of-the-art segmentation performance on three different (synthetic, over-the-air, and large-scale real-world) datasets, outperforming leading RVNN models.
    \item We provide the dataset on IEEE DataPort to support reproducibility and further research: \url{https://dx.doi.org/10.21227/hcc1-6p22}.
\end{itemize}
The remainder of this paper is organized as follows: 
Related work on spectrum segmentation and CVNN applications is discussed in Section~\ref{sec:related}. Background on CVNNs is provided in Section~\ref{sec:background}. The problem definition and the $\mathbb{C}$MuSeNet architecture, including a novel complex-valued loss function, are discussed in Section ~\ref{sec:segment}. The hyperparameter analysis and comparative evaluation results are shown in Section~\ref{sec:eval}. Finally, conclusions and potential directions for future research are discussed in Section~\ref{sec:conclusion}.

\section{Related Work}
\label{sec:related}
Spectrum segmentation is essential for identifying available segments within a wideband environment in dynamic spectrum sharing. However, this task is made challenging by low-SNR conditions and multiple signals~\cite{Subedi24, Oshea16endtoend, Huang20detection_fcn, Lin22detection_pyramid}. RVNNs have been widely applied in this domain, improving segmentation accuracy for spectrum sensing. 
\vspace{-0.03in}
\subsection{Real-Valued Neural Networks in Spectrum Segmentation}
\vspace{-0.01in}
RVNNs primarily process power spectral density (PSD) data to enhance spectrum segmentation. For instance, a Fully Convolutional Network (FCN)~\cite{Huang20detection_fcn} detects carrier signals in broadband spectra. The DeepMorse framework~\cite{Yuan19detection_morse} bypasses threshold-based methods by directly extracting features from PSD data for signal detection. Similarly, the Seek and Classify framework~\cite{Subedi24} integrates segmentation and modulation classification, improving accuracy in real-time cognitive radio applications. In ~\cite{Li22ResNet22}, ResNet-22 is employed for spectrum segmentation, demonstrating RVNN performance for this task. However, these methods rely on real-valued inputs, falling short of capturing critical in-phase and quadrature (IQ) signal properties, especially in low-SNR conditions. 
\vspace{-0.05in}
\subsection{Complex-Valued Neural Networks}
\vspace{-0.01in}
CVNNs address RVNN limitations by preserving phase and amplitude information, making them effective for complex-valued data processing~\cite{hirose2012complexnn}. Deep Waveform~\cite{Zhao21DeepWaveform}, a CVNN-based OFDM receiver, outperforms legacy OFDM receivers under fading channels by retaining signal structure. 

The limitations of real-valued backpropagation for CVNNs motivate the need for specialized training methods to leverage CVNN capabilities fully~\cite{Tan22RVBPnotSuitableforCVNN}. Key architectural components, such as complex convolutions, activation functions, and batch normalization, form the foundation of effective CVNN designs~\cite{trabelsi2018deepnn}. Recent work has demonstrated the utility of CVNNs in high-fidelity applications, such as MRI fingerprinting~\cite{Virtue17CVNNMRI}, synthetic aperture radar (SAR) classification~\cite{Scarnati21CVNNSAR}, and radio signal recognition~\cite{Xu22AnalysisCVNN_RF}. These studies showcase CVNNs' abilities to preserve phase and amplitude information of periodic signals, achieving superior performance under challenging conditions such as the low-SNR regime.

Despite their success, the application of CVNNs to spectrum segmentation remains largely unexplored. Prior spectrum segmentation results have typically relied on real-valued networks, resulting in low accuracy for weak signals. In this work, we design a complex-valued neural network specifically for spectrum segmentation, processing IQ signals to retain critical signal characteristics. This capability is particularly beneficial in low-SNR conditions. Furthermore, we introduce novel complex-valued loss functions and metrics for broad applicability across CVNN tasks. This contribution advances both spectrum sensing and CVNN research.
\begin{figure}[t!]
    \centering
    \includegraphics[width=0.27\textwidth]{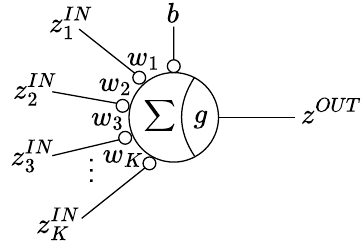}
    \caption{Complex-valued neuron~\cite{hirose2012complexnn}.}
    \label{fig:CVNeuron}\vspace{-0.3in}
\end{figure} 
\vspace{-0.02in}
\section{Background}
\label{sec:background}
\vspace{-0.01in}
\subsection{Complex-Valued Neural Networks}
CVNNs operate within the complex domain, enabling direct processing of IQ signals as well as complex-valued Fourier transform by retaining phase and amplitude information. A complex-valued neuron extends the real-valued neuron, designed to process inputs, weights, and output in the complex domain, as shown in Fig.~\ref{fig:CVNeuron}. The complex-valued neuron is defined as~\cite{hirose2012complexnn}:
\begin{equation}\label{equation:complex-neuron}
    z^{OUT} = g\left(\sum_{k=1}^Kw_kz_k^{IN} + b\right),
\end{equation}
where $z^{IN}_k$ is the $k$-th complex-valued input and $z^{OUT}$ is the complex-valued output, $g$ is the activation function, $w_k$ is complex-valued weights, and $b$ is the bias.

To train CVNNs, backpropagation is adapted to accommodate complex-valued weights~\cite{hirose2012complexnn}. This is achieved through Wirtinger derivatives, which enable differentiation with respect to both the real and imaginary components independently~\cite{Zhao21DeepWaveform}. 
For a complex-valued weight $w_k = w_{x,k} + jw_{y,k}$ where $j=\sqrt{-1}$, $w_{x,k}$ and $w_{y,k}$ are the real and imaginary component of the weight respectively, the gradient of a loss function $L$ with respect to $w_k$ is given by~\cite{kreutzdelgado2009cvnnGradient}: 
\begin{equation} \label{equation:WritingerDerivation}
    \frac{\partial L}{\partial w_k} = \frac{\partial L}{\partial w_{x,k}} + j\frac{\partial L}{\partial w_{y,k}}. \vspace{-0.05in}
\end{equation}
This formulation allows independent weight updates in real and imaginary parts, preserving the complex-valued structure throughout training. 

For convolutional operations in CVNNs, a complex-valued convolutional filter $W = A + jB$ is used, where $A$ and $B$ represent the real and imaginary components, respectively. This filter operates on complex-valued inputs $z = x + jy$ and the output of the convolution is given by~\cite{Xu22AnalysisCVNN_RF, trabelsi2018deepnn}:
\begin{equation} \label{equation:CVConvolution}
    (W \ast z) = (A \ast x - B \ast y) + j(B \ast x + A \ast y), 
\end{equation} 
where $*$ denotes the convolution operation. Such operations are critical for implementing deep learning structures capable of processing IQ data in its native complex form.

\subsection{Transformations and Activation for CVNNs}
In this section, we describe the key building blocks for CVNNs, including activation functions, transformations, and batch normalization. In CVNNs, activation and transformation functions are applied independently to real and imaginary components~\cite{Lee22CVNNSurvey}:\vspace{-0.05in}
\begin{equation}\label{equation:CVSplitfunction}\vspace{-0.05in}
    g(z) = g(x) + jg(y)
\end{equation}
where $g(x)$ and $g(y)$ represent the function applied to the respective components.
$\mathbb{C}$ReLU utilizes this characteristic and defined as~\cite{trabelsi2018deepnn}:
\begin{equation}\label{equation:CVReLu}
    \mathbb{C}ReLU(z) = ReLU(x) + jReLU(y).
\end{equation}
Holistic activation functions, in contrast, leverage the magnitude and phase of the input as a unified representation. For instance, modReLU~\cite{trabelsi2018deepnn} adjusts the magnitude while preserving the phase of the complex input. 
Split-based and holistic approaches provide complementary methodologies for designing CVNN activation functions. In this work, $\mathbb{C}$ReLU is employed for its efficiency and compatibility with the proposed CVNN architecture. 
Transformations, such as $\mathbb{C}$sigmoid and $\mathbb{C}$Average pooling, also follow the split-based approach, defined as~\cite{trabelsi2018deepnn}:
\begin{equation}\label{equation:CVSigAvgPool}
\begin{split}
    \mathbb{C}sigmoid(z) &= sigmoid(x) + jsigmoid(y), \\
    \mathbb{C}Avg.Pool(z) &= Avg.Pool(x) + jAvg.Pool(y).
\end{split}
\end{equation}
While activation and transformation functions address non-linearity, linear and batch normalization transformations operate directly on complex numbers. 
The complex linear transformation layer is defined as~\cite{Xu22AnalysisCVNN_RF}:
\begin{equation} \label{equation:CVLinear} 
w_kz + b = (w_{x,k} \cdot x - w_{y,k} \cdot y + b_x) + j(w_{y,k} \cdot x + w_{x,k} \cdot y + b_y), 
\end{equation}
where $w_{x,k}$ and $w_{y,k}$ denote the real and imaginary components of the $k$-th complex-valued weights, and $b=b_x+jb_y$ denotes the complex bias.
The complex-valued batch normalization is defined as~\cite{trabelsi2018deepnn}:
\begin{equation}\label{equation:CVBN}
    \mathbb{C}BN(z) = \gamma_z \cdot z_{norm} + \beta_z,
\end{equation}
where $\gamma_z = \gamma_x+j\gamma_y$ is a scaling parameter to adjust the normalized input to match the desired variance, $\beta_z = \beta_x+j\beta_y$ is the shifting parameter to center the output to a target mean. The $z_{norm}$ is the normalization of $z$ given by~\cite{trabelsi2018deepnn}:
\begin{equation}\label{equation:CVBNNormalization}
    z_{norm} = V^{-1/2}\cdot(z - \mu_z),
\end{equation}
where $\mu_z = \mu_x + j\mu_y$ is the mean of the complex-valued batch, 
and V is the covariance matrix of $x$ and $y$:
\begin{equation}\label{equation:CVBNVMatrix}
    V =
        \begin{bmatrix}
        \sigma_x^2 & \text{Cov}(x, y) \\
        \text{Cov}(y, x) & \sigma_y^2
        \end{bmatrix},
\end{equation}
with $\sigma_x^2$ and $\sigma_y^2$ as variance, and Cov(x,y) as the covariance.

While complex-valued activation and transformation functions enable CVNNs to process complex-valued data effectively, traditional loss functions and evaluation metrics are typically defined for real-valued outputs. These conventional methods are not suitable for CVNNs that produce complex-valued outputs, as they fail to capture the relationships inherent in complex data. Consequently, there is a need to develop specialized loss functions and validation metrics that can handle complex-valued outputs, ensuring effective training and evaluation of CVNNs in spectrum segmentation tasks. 

\section{Complex-Valued Multi-Signal Segmentation}
\label{sec:segment}
\subsection{Problem Definition}
Consider a wideband spectrum sensor that monitors the spectrum to detect spectrum occupation and spectrum gaps by detecting multiple signals. The received wideband signal, $r(t)$, is modeled as the sum of $N$ signals: 
\begin{equation}\label{equation:AWGNMultisig}
    r(t) = \sum_{i=1}^N s_i(t) + n(t),
\end{equation}
where $s_i(t)=\Re\{s_{c,i}(t)e^{j2\pi f_{c,i}t}\}$ represents the $i$-th signal, $s_{c,i}(t)$ and $f_{c,i}$ are the baseband complex envelope and the carrier frequency of the $i$-th signal, and $n(t)$ represents the additive white Gaussian noise (AWGN) with zero mean and a variance of $\sigma^2$. 

To sample this signal, assume that the sensor mixes the signal with a center frequency, $f_r$, with a sampling rate of $f_s=1/T_s$ and applies low-pass filtering that retains signals with $|f_{c,i}-f_r|\leq f_s/2$. This leads to  a generalized form of discrete-time in-phase and quadrature (IQ) samples:
\begin{equation}\label{equation:AWGNDiscreteMultisig}
    r_{IQ}[n] = \sum_{i:|f_{c,i}-f_r|\leq f_s/2} s_{c,i}[n]e^{j2\pi (f_{c,i}-f_r)nT_s}\} + n_c[n],
\end{equation}
where $r_{IQ}[n]=r_I[n]+jr_Q[n]$ is the complex-valued sample representation. 

During a sampling window of $\tau_s$, the receiver collects $L = \tau_s \times f_s$ discrete IQ samples. Transforming these samples into the frequency domain using the Discrete Fourier Transform (DFT) yields:
\begin{equation}\label{equation:dft}
    R[f] = \sum_{n=0}^{L-1}r_{IQ}[n]e^{-j2\pi fn/L},\; f \in \{0,1,\ldots , L-1\},
\end{equation}
where $R[f]=R_x[f] + jR_y[f]$ represents the complex-valued Fourier transform (FT) coefficient at the $f$-th frequency bin, $r_{IQ}[n]$ denotes the $n$-th IQ sample, and $L$ is the number of DFT points. Now, we can define the \textbf{spectrum segmentation problem}: Our goal is to determine the set of start and end frequency ranges for each occupied segment, in which significant energy is present, based on $R[f]$: $\mathcal{B} = \{ (f_i^b, f_i^e) \, : \, \forall i \in \{1, \dots, N\} \}$, 
where $f_i^b$ and $f_i^e$ denote the beginning and end points of the $i$-th occupied segment respectively, and $N$ is the total number of occupied segments. The objective is to determine the set $\mathcal{B}$, identifying the boundaries of all occupied frequency segments within the spectrum. The problem is illustrated in Fig.~\ref{fig:spec_seg_illust}, where raw IQ samples (left) are used to segment the spectrum showing occupied bands (right).

Next, we present the $\mathbb{C}$MuSeNet architecture as well as the novel architectural and training components to address this problem.

\begin{figure}[t!]
    \centering
    \includegraphics[width=0.4\textwidth]{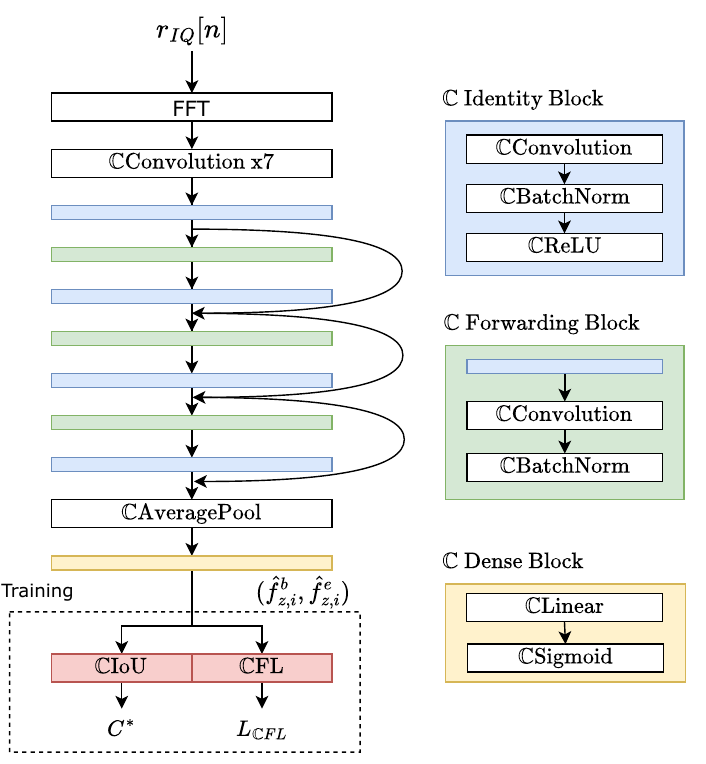}\vspace{-0.05in}
    \caption{$\mathbb{C}$MuSeNet architecture.}
    \label{fig:CV-ResNet}\vspace{-0.3in}
\end{figure}
\subsection{$\mathbb{C}$MuSeNet Architecture}
$\mathbb{C}$MuSeNet is designed for spectrum segmentation in dynamic spectrum access systems.
The $\mathbb{C}$MuSeNet architecture with a training component is shown in Fig.~\ref{fig:CV-ResNet}. It employs a residual network structure that effectively captures complex-valued IQ signal characteristics while addressing the vanishing gradient problem in deep networks.

In $\mathbb{C}$MuSeNet, the complex-valued input $r_{IQ}[n]$ represents the wideband IQ samples containing multiple signals the model aims to segment. This input undergoes a Fast Fourier Transform (FFT) to convert the time-domain signal into the frequency domain while maintaining its complex-valued nature. The FFT transformation aligns the data with the frequency components essential for spectrum segmentation tasks. 

The $\mathbb{C}$MuSeNet architecture (Fig.~\ref{fig:CV-ResNet}) includes seven $\mathbb{C}$Convolutional layers, designed to extract low-level features while preserving both magnitude and phase information as defined in (\ref{equation:CVConvolution}). These initial layers are followed by multiple $\mathbb{C}$ residual blocks, each comprising $\mathbb{C}$ identity and $\mathbb{C}$ forwarding blocks. Each residual block contains $\mathbb{C}$Convolutional layers, $\mathbb{C}$Batch normalization, and $\mathbb{C}$ReLU activation functions. Including identity mapping in each residual block facilitates residual learning, effectively mitigating the vanishing gradient problem and ensuring robust feature extraction for complex-valued IQ signals.

Including $\mathbb{C}$Batch normalization within the identity blocks normalizes activations in the complex-valued domain, enhancing training stability and convergence speed. This normalization addresses internal covariate shifts in complex-valued weights and activations, improving the training process.
Additionally, applying $\mathbb{C}$ReLU within these blocks enables the network to focus on relevant features while suppressing irrelevant ones, facilitating the learning of complex-valued representations.

Following the identity and forwarding blocks, a $\mathbb{C}$Average pooling layer reduces spatial dimensions while aggregating learned features. This is followed by a $\mathbb{C}$Dense block comprising a $\mathbb{C}$Linear layer and a $\mathbb{C}$Sigmoid function, which is applied separately to the real and imaginary components, as defined in (\ref{equation:CVLinear}). This approach allows the model to effectively identify the presence or absence of signals within specific frequency ranges, enabling precise binary spectrum segmentation.

While the architecture forms a crucial aspect of $\mathbb{C}$MuSeNet, its primary contributions lie in the specialized training components we introduce. Traditional loss functions and training evaluation metrics are inadequate for complex-valued outputs, as they do not account for the relationships in complex-valued data. To address this, we develop novel complex-valued training elements as described next.

\subsection{Complex-Valued Fourier Spectrum Loss Function}
To leverage the complex-valued Fourier spectrum provided as an input to $\mathbb{C}$MuSeNet, we propose a new complex-valued loss function for spectrum segmentation: Complex-Valued Focal Loss ($\mathbb{C}$FL). In RVNNs, focal loss has been utilized in binary classification tasks, which addresses class imbalance and applies a modulating factor to focus more on hard-to-classify examples. This feature is desirable for low-SNR signal segmentation, but complex-valued counterparts of FL do not exist. To this end, the proposed $\mathbb{C}$FL is represented as:
 \begin{equation} \label{equation:CVFL}
    \begin{split}
        L_\text{$\mathbb{C}$FL} = 
        & - \frac{\alpha}{2} \sum_{f=0}^{L-1} \bigg((1 - p_{x,f})^\gamma o_{x,f} \log(p_{x,f}) \\
        & + (p_{x,f})^\gamma (1 - o_{x,f}) \log(1 - p_{x,f}) \\
        & + (1 - p_{y,f})^\gamma o_{y,f} \log(p_{y,f}) \\
        & + (p_{y,f})^\gamma (1 - o_{y,f}) \log(1 - p_{y,f})\bigg),
    \end{split}
 \end{equation}
where $\alpha$ is the weighting factor; $L$ is the number of frequency bins; $\gamma$ is the focusing parameter; $o_{x,f}$ and $o_{y,f}$ are the binary ground truths indicating whether a signal occupies a frequency bin $f$ based on the real and imaginary parts of the FT coefficient, respectively; and $p_{x,f}$ and $p_{y,f}$ are the corresponding predicted probabilities. Accordingly, we convert signal occupancy at a frequency bin into a binary classification problem. The factor of $1/2$ ensures that the total loss balances the contribution of both real and imaginary parts.

$\mathbb{C}$FL is inspired by the fact that the Fourier transform of a signal (or a number of signals) generally results in non-zero real and imaginary values at the frequency bin they occupy\footnote{We acknowledge rare cases where an FT coefficient lies on the real or imaginary axis, which we disregard due to their negligible impact.}. Accordingly, the real and imaginary components of the FT are treated as two separate predictions. This allows the loss function to retain phase information, which is lost if the magnitude spectrum is considered instead. Through the focusing parameter, $\gamma$, contributions of easy-to-detect bins are modulated to focus learning on hard-to-detect frequency bins. This approach enhances learning performance in complex-valued tasks. In Section~\ref{sec:eval}, we evaluate $\mathbb{C}$FL with respect to a complex-valued version of the binary cross-entropy loss function and fine-tune the weighting factor and focusing parameter to improve performance. $\mathbb{C}$FL is used to update the neural network weights based on (\ref{equation:WritingerDerivation}) as well as a part of the stopping criterion, as described next.

\subsection{Complex Plane Intersection Over Union Similarity Metric}
To improve the training process for spectrum segmentation, we propose an extended boundary-based IoU similarity metric specifically designed for complex-valued signals. $\mathbb{C}$IoU is used as the second part of the stopping criterion during training. This approach avoids reducing the FT into real-valued metrics such as the magnitude spectrum, thereby preserving the critical phase information for accurately capturing the boundaries of signals in the frequency domain.

In this method, we extend the spectrum segmentation problem from a one-dimensional segment to a \textbf{two dimensional area} where the occupied segments in each dimension are based on the real and imaginary components of the FT coefficients. For a signal $i$, let the target area be $B_{z,i} = \{ (x, y) \mid f^b_{x,i} \leq x \leq f^e_{x,i} \text{ and } f^b_{y,i} \leq y \leq f^e_{y,i} \}$ based on the beginning and ending frequency bins of a signal according to the real and imaginary components of the FT coefficients. Similarly, the predicted area is represented as $\hat{B}_{z,j} = \{ (x, y) \mid \hat{f}^b_{x,j} \leq x \leq \hat{f}^e_{x,j} \text{ and } \hat{f}^b_{y,j} \leq y \leq \hat{f}^e_{y,j} \}$. The complex plane IoU measures the similarity between predicted and ground truth areas and is defined as:  
\begin{equation} \label{equation:CVIoU}
\mathbb{C}IoU_{i,j} = \mathcal{A}(B_{z,i} \cap \hat{B}_{z,j})/\mathcal{A}(B_{z,i} \cup \hat{B}_{z,j}),
\end{equation}
where $\mathcal{A}(\cdot)$ is the area operator. Given a set of $N$ target boundaries, $\mathcal{B}_z = \{B_{z,i} \ | \ i = 1 ,...,N\}$ and a set of $\hat{N}$ predicted boundaries $\hat{\mathcal{B}}_z = \{\hat{B}_{z,j} \ | \ j = 1, ... , \hat{N}\}$, the IoU matrix is denoted as $\mathbf{C}\in \mathbb{R}^{Nx\hat{N}}$, where each element is given by $C_{i,j} = \mathbb{C}IoU_{i.j}$.
The optimal matching maximizes the total IoU with the condition of a one-to-one assignment: 
\begin{equation} \label{equation:CVIoUMax}
\mathbf{C}^* = \arg\max_{\mathbf{X}\in \{0,1\}^{Nx\hat{N}}} \sum_{i,j} X_{i,j} \mathbb{C}IoU_{i,j}
\end{equation}
where $X_{i,j} \in \mathbf{X}$ is the binary decision variable and $\mathbf{C}^*$ represents the optimal binary assignment matrix.
Along with the loss function in (\ref{equation:CVFL}), $\mathbb{C}IoU_{i,j}$ is used as a stopping criterion, where the training is stopped if the loss and the IoU score do not improve above a certain threshold. This approach leads to improved accuracy and better segmentation performance compared to converting complex FFT results into real-valued forms, as discussed in Section~\ref{sec:eval}. 


\section{Performance Evaluations}
\label{sec:eval}
The $\mathbb{C}$MuSeNet framework is evaluated on three datasets: a synthetic dataset with additive white Gaussian noise (AWGN) channel, an indoor over-the-air (OTA) dataset reflecting realistic transmission scenarios, and the broadband irregularly-sampled geographical radio environment dataset (BIG-RED), comprising extensive real-world spectrum data. This diverse evaluation setup highlights the framework's effectiveness across varying environments and data complexities. Table ~\ref{table:dataset-parameters} summarizes the dataset parameters.
\subsection{Datasets and Evaluation Setup}
\subsubsection{Synthetic Dataset}
The synthetic dataset is created using MATLAB, generating wideband IQ samples containing multiple signals designed to cover diverse scenarios. Signals are generated with various modulation types, including BPSK, QPSK, 8-PSK, 8-QAM, 16-QAM, GMSK, and 2-FSK. Each signal is placed within the receiver's frequency range, with a guard band of $0.1$ MHz between signals to avoid overlap in the frequency domain. 
To simulate varying noise levels, AWGN is introduced to achieve SNR values ranging from $-20$ dB to $10$ dB. The noise power is distributed uniformly across the entire sample bandwidth of $20$ MHz. Samples are evenly distributed across the SNR range, ensuring comprehensive coverage of noise conditions for evaluation in wideband signal segmentation tasks.
\begin{figure}[t!]
    \centering
    \begin{subfigure}[b]{0.24\textwidth}
        \centering
        \includegraphics[width=\textwidth]{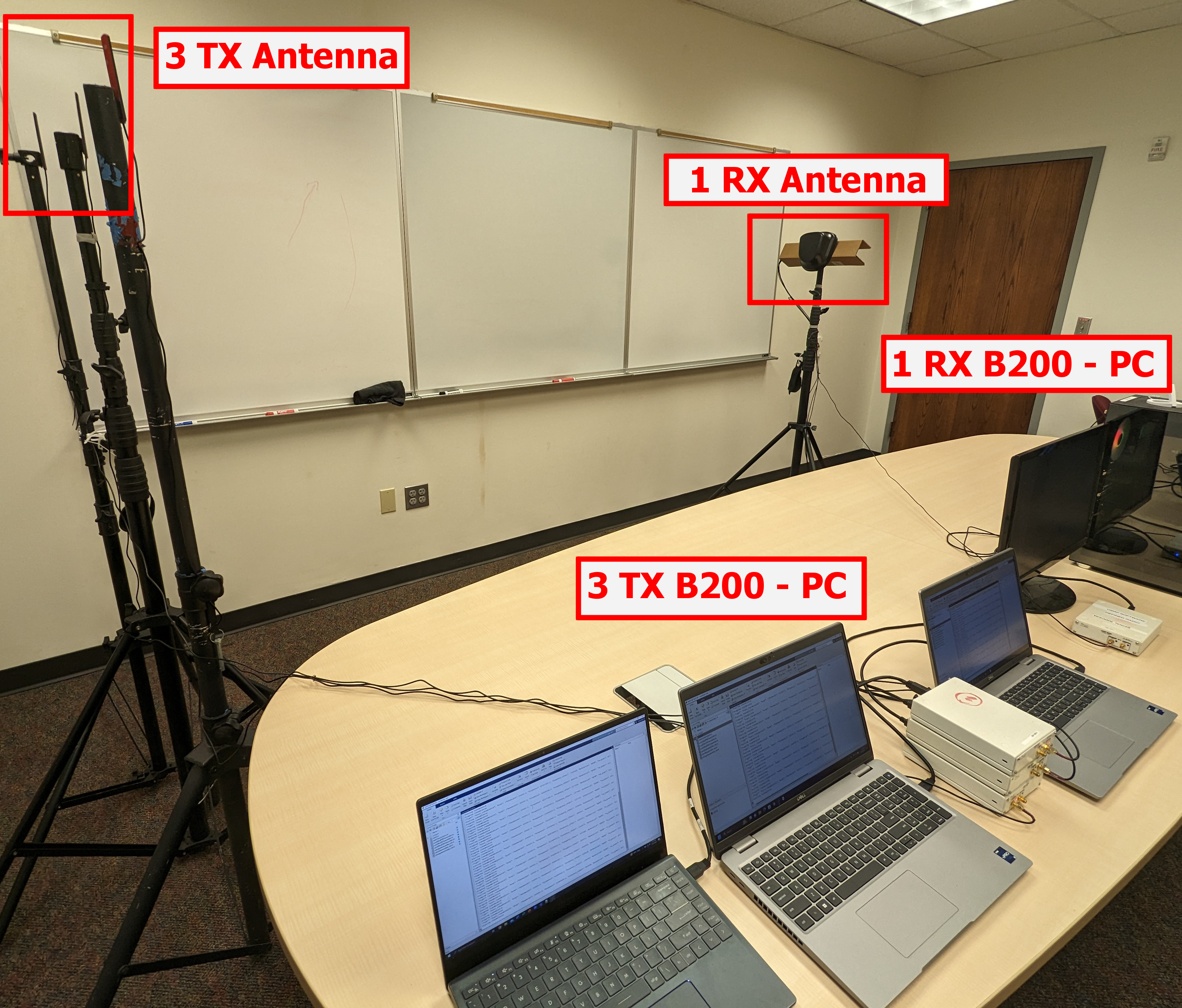}
        \caption{OTA dataset testbed setup}
        \label{fig:OTA-Setup}
    \end{subfigure}
    \hfill
    \begin{subfigure}[b]{0.24\textwidth}
        \centering
        \includegraphics[width=\textwidth]{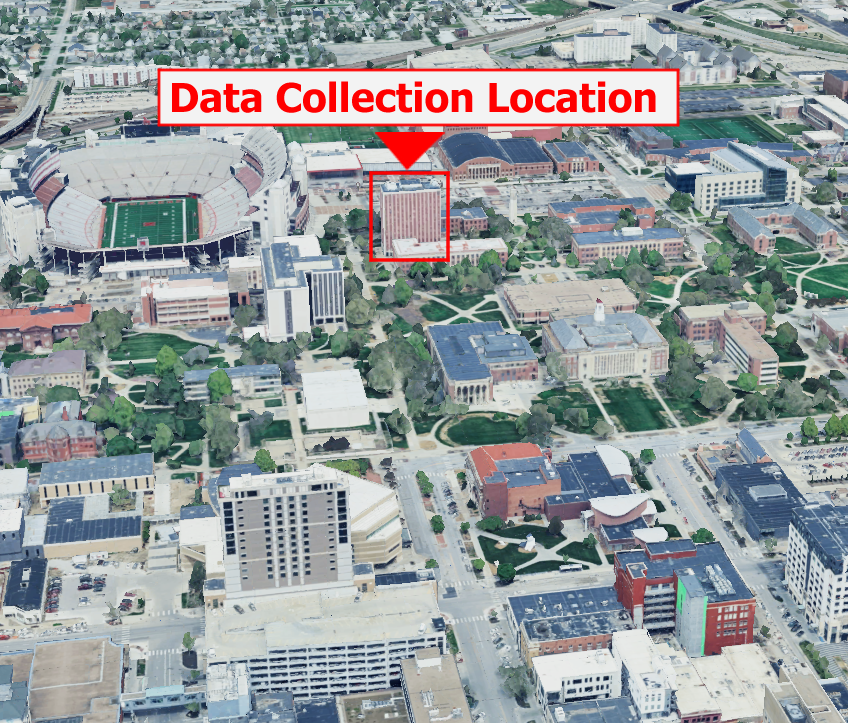}
        \caption{BIG-RED data collection site}
        \label{fig:BIG-REDCollection}
    \end{subfigure} \vspace{-0.25in}
    \caption{Dataset collection setups for OTA and BIG-RED}
    \label{fig:merged-figures}
    \vspace{-0.2in}
\end{figure}
\subsubsection{Indoor OTA Dataset}
The Indoor OTA dataset~\cite{Subedi24} is collected using a software-defined radio (SDR) testbed, as shown in Fig.~\ref{fig:OTA-Setup}. The testbed includes three Ettus USRP B200 transmitters and one Ettus USRP B200 receiver, each equipped with a sub-6 GHz wideband antenna. The transmitter randomly selects modulation types from BPSK, QPSK, and 2-FSK. Signals are transmitted in the $900-920$ MHz Industrial, Scientific, and Medical (ISM) band, with bandwidths ranging from $0.1$ MHz to $2$ MHz. The receiver captures the transmitted signals at a sampling rate of $20$ MHz. This testbed introduces realistic channel effects such as interference and multipath propagation while maintaining controlled conditions. Noise power is measured from samples collected when the transmitters are off, while signal power is calculated from segmented signals. Using this information, sample SNR is determined to quantify the performance of the $\mathbb{C}$MuSeNet framework.
\subsubsection{BIG-RED}
BIG-RED is a real-world spectrum dataset collected since $2020$ using the Nebraska Experimental Testbed of Things (NEXTT)~\cite{ZHAO21NEXTT}, a city-wide distributed outdoor wireless experimental testbed composed of Ettus USRP N310 SDRs. BIG-RED spans frequencies from $54$ MHz to $2.6$ GHz, capturing diverse signal conditions, dynamic channel usage, environmental noise, and interference patterns.
The complete BIG-RED comprises over $123$ TB of wideband IQ samples. 
For evaluation, a subset of $10,000$ samples ($1.7$ TB) from one location is processed, with IQ sample durations reduced by a factor of 10 for efficient GPU utilization, resulting in a $34$ GB dataset. This subset retains the diversity of the original dataset, which includes signals with varying modulation types, strengths, and bandwidths. 
\begin{table}[t!]
\centering
\caption{Dataset Parameters}
\begin{adjustbox}{width=0.4\textwidth}
{
\begin{tabular}{|l|c|c|c|} \hline
 & Synthetic & Indoor OTA & BIG-RED\\ \hline
 No.~of signals, \(N\) & 1 -- 10 & 1 -- 3 & 1 -- 9 \\ \hline
 Sampling rate, $f_s$ & \multicolumn{3}{c|}{20 MHz} \\ \hline
 Sample SNR (dB) & (-20) -- 10 & (-10) -- 10 & - \\ \hline
 Signal BW (MHz) & \multicolumn{2}{c|}{0.1, 0.2, 0.5, 1, 2} & 0.12 - 10.75 \\ \hline
 No.~of samples used & 80,000 & 18,600 & 10,000 \\ \hline
\end{tabular}
}
\end{adjustbox}
\label{table:dataset-parameters}\vspace{-0.25in}
\end{table}
Ground truth is derived by converting IQ samples to the frequency domain to verify transmissions. Unlike synthetic and OTA datasets, BIG-RED lacks predefined signal constraints, making it particularly challenging. Evaluation is performed across the dataset to assess $\mathbb{C}$MuSeNet's robustness in dynamic and unpredictable real-world spectral environments.
Due to the complexity and diversity of the BIG-RED, evaluation is performed across the entire dataset rather than per SNR, emphasizing the robustness of the $\mathbb{C}$MuSeNet in dynamic and unpredictable real-world spectral environments.
\subsubsection{Evaluation Methodology}
The evaluations are conducted on a node equipped with two Intel Xeon Silver 4110 CPUs, an NVIDIA Tesla V100 GPU with $16$ GB of VRAM, and $187$ GB of RAM, running Ubuntu 18.04. Models are implemented in Python 3.10 using PyTorch~\cite{pytorch} and ComplexPyTorch~\cite{Sebastien23complexPyLib} libraries. Each dataset is divided into training, validation, and testing sets using an 80\%-10\%-10\% split.  Training is performed until validation accuracy does not improve and validation loss does not decrease for three consecutive epochs. The learning rate is initialized at $0.001$, and reduces to $0.0001$ if training metrics stagnate for three epochs. These parameters are selected through hyperparameter analysis (Section~\ref{subsec:hyper}), optimizing validation accuracy and minimizing training epochs.
Transfer learning is employed to handle datasets with varying complexities. The model is first trained on the synthetic dataset to establish robust low-SNR performance. This pre-trained model is subsequently fine-tuned on the indoor OTA dataset and then, BIG-RED, allowing it to retain low-SNR capabilities while adapting to specific dataset characteristics. This sequential fine-tuning strategy enhances both accuracy and efficiency, outperforming models trained from scratch~\cite{Zhao21DeepWaveform,Subedi24}.

\begin{figure*}[t!]
\centering
\begin{subfigure}{0.50\columnwidth}
  \centering
  \includegraphics[width=\textwidth]{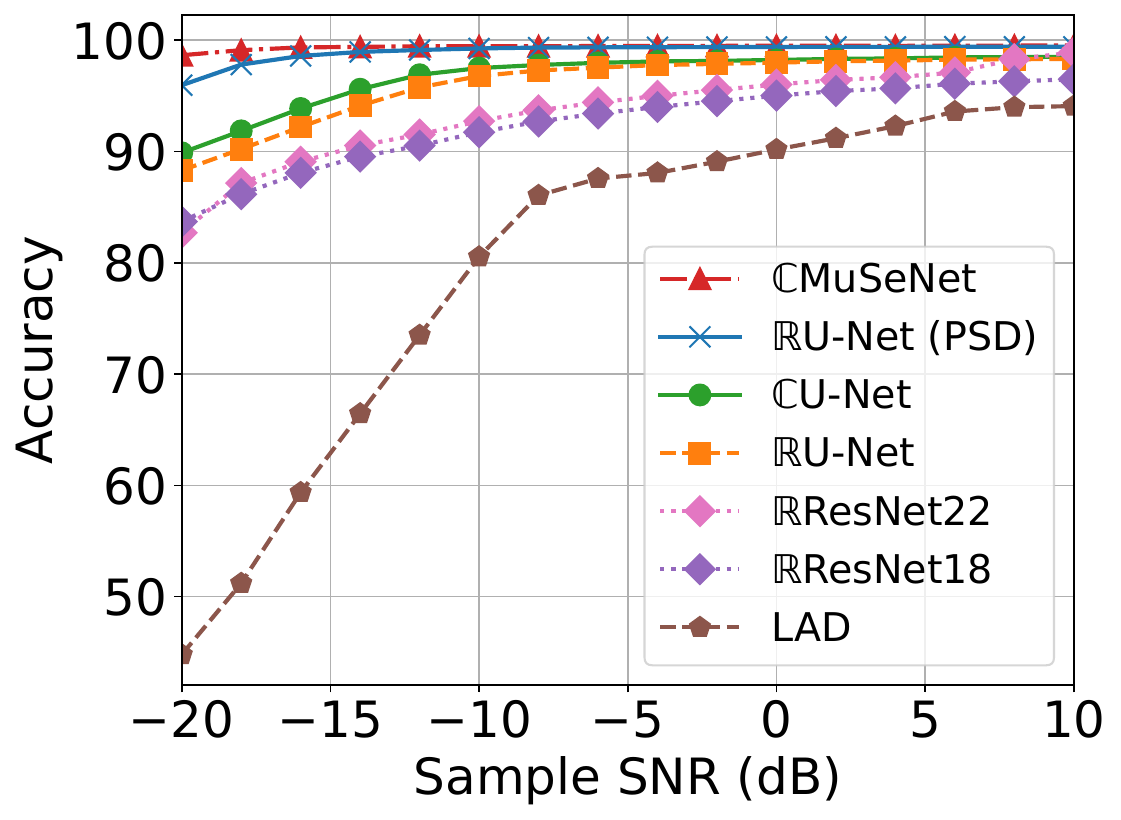}
  \caption{Accuracy}
  \label{fig:Syn_Accuracy}
\end{subfigure}%
\begin{subfigure}[b]{0.51\columnwidth}
  \centering
  \includegraphics[width=\textwidth]{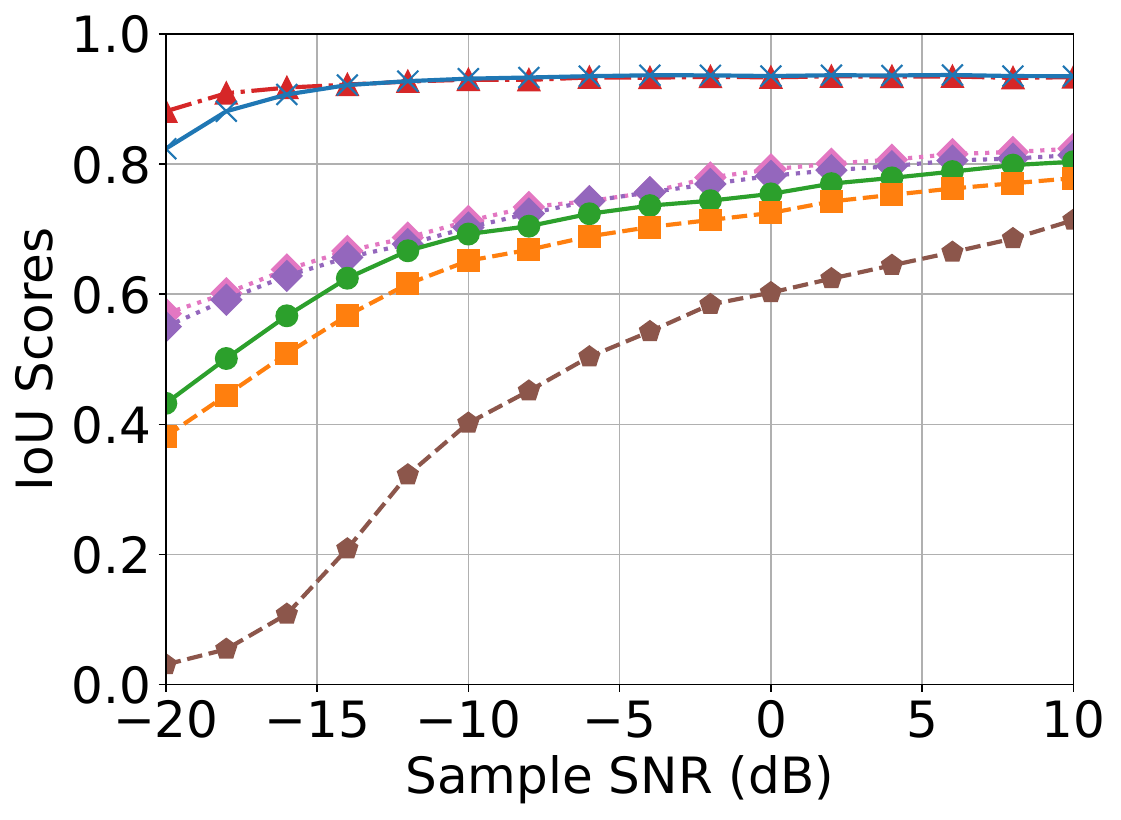}
  \caption{IOU Score}
  \label{fig:Syn_IoU}
\end{subfigure}%
\begin{subfigure}[b]{0.51\columnwidth}
  \centering
  \includegraphics[width=\textwidth]{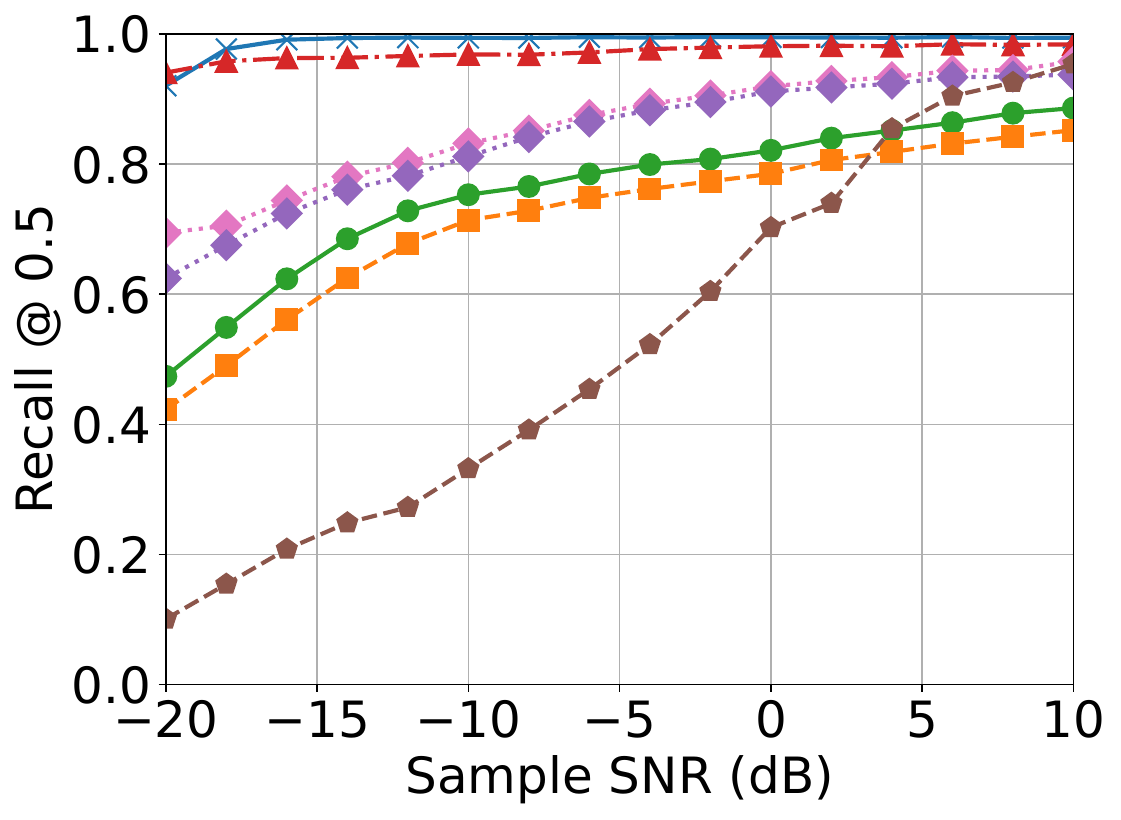}
  \caption{Recall (threshold=0.5)}
  \label{fig:Syn_Recall_0.5}
\end{subfigure}%
\begin{subfigure}[b]{0.51\columnwidth}
  \centering
  \includegraphics[width=\textwidth]{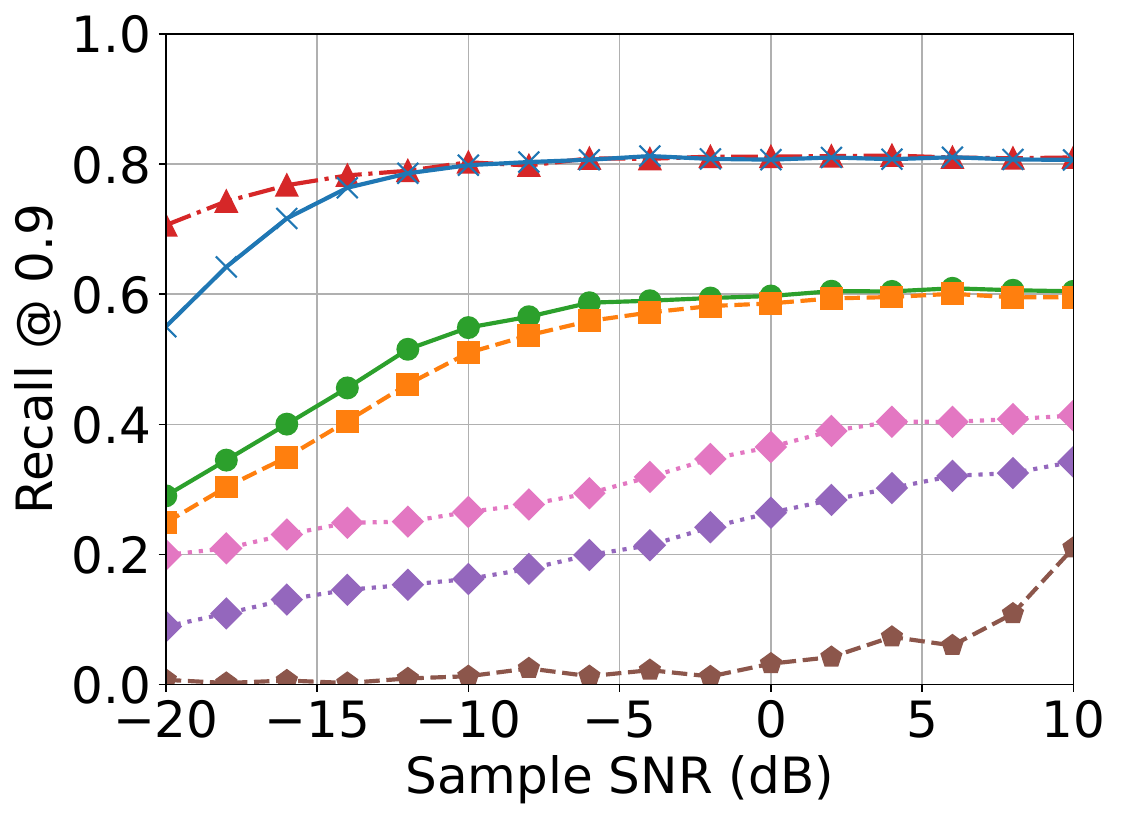}
  \caption{Recall (threshold=0.90)}
  \label{fig:Syn_Recall_0.9}
\end{subfigure}%
\vspace{-0.05in}
\caption{Synthetic Dataset Spectrum segmentation performance under different SNR conditions.}
\label{fig:Syn_Performance}\vspace{-0.2in}
\end{figure*}
\begin{figure*}[t!]
\centering
\begin{subfigure}{0.51\columnwidth}
  \centering
  \includegraphics[width=\textwidth]{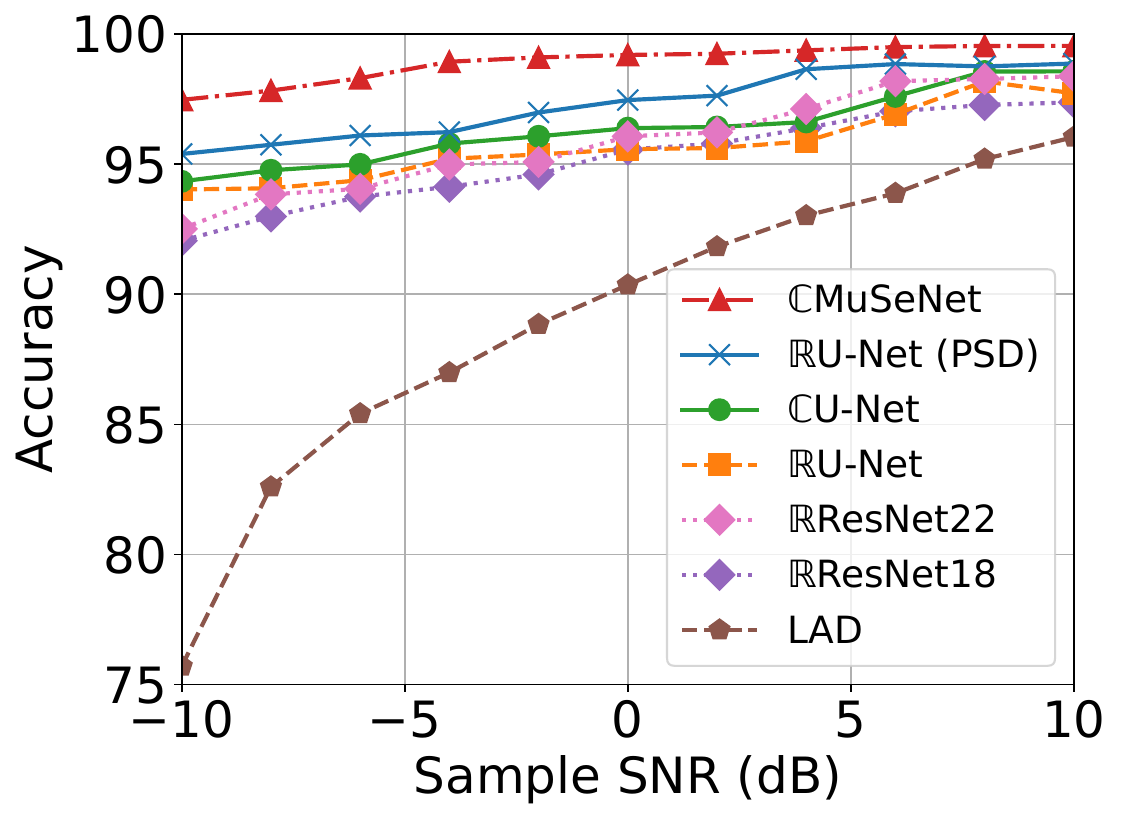}
  \caption{Accuracy}
  \label{fig:OTA_Accuracy}
\end{subfigure}%
\begin{subfigure}[b]{0.51\columnwidth}
  \centering
  \includegraphics[width=\textwidth]{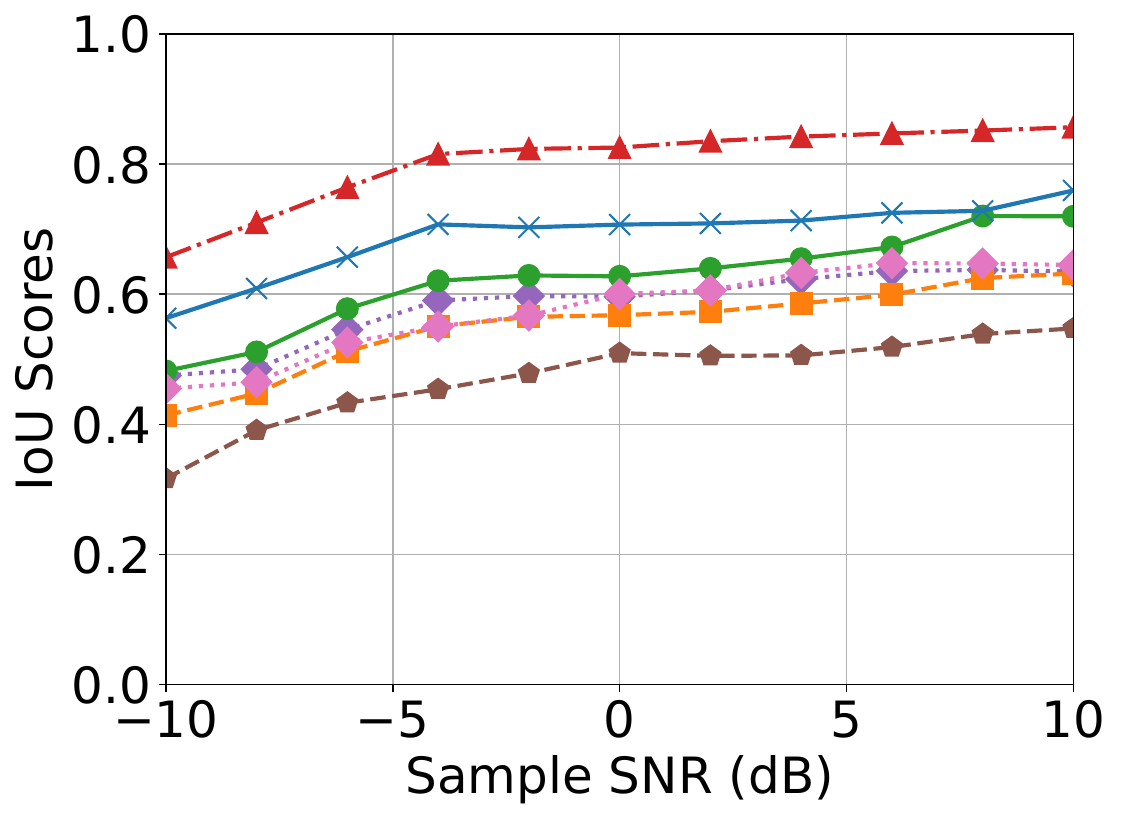}
  \caption{IOU Score}
  \label{fig:OTA_IoU}
\end{subfigure}%
\begin{subfigure}[b]{0.51\columnwidth}
  \centering
  \includegraphics[width=\textwidth]{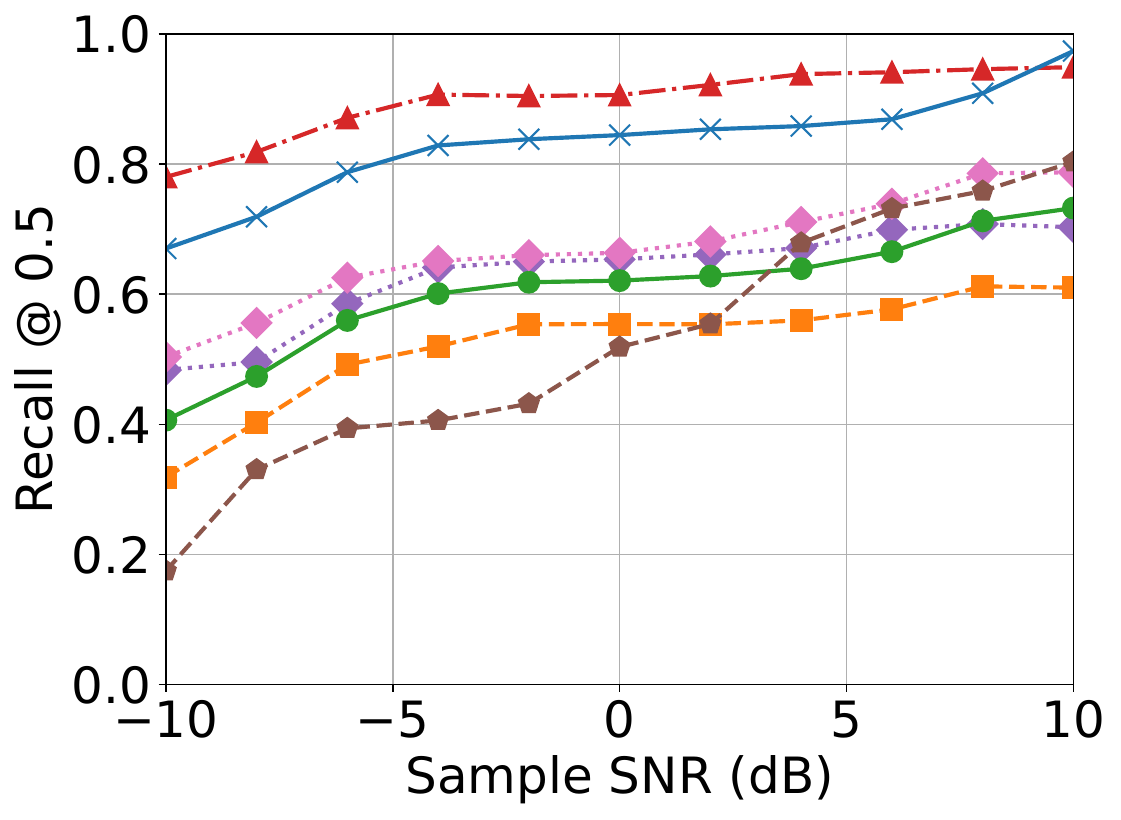}
  \caption{Recall (threshold=0.5)}
  \label{fig:OTA_Recall_0.5}
\end{subfigure}%
\begin{subfigure}[b]{0.51\columnwidth}
  \centering
  \includegraphics[width=\textwidth]{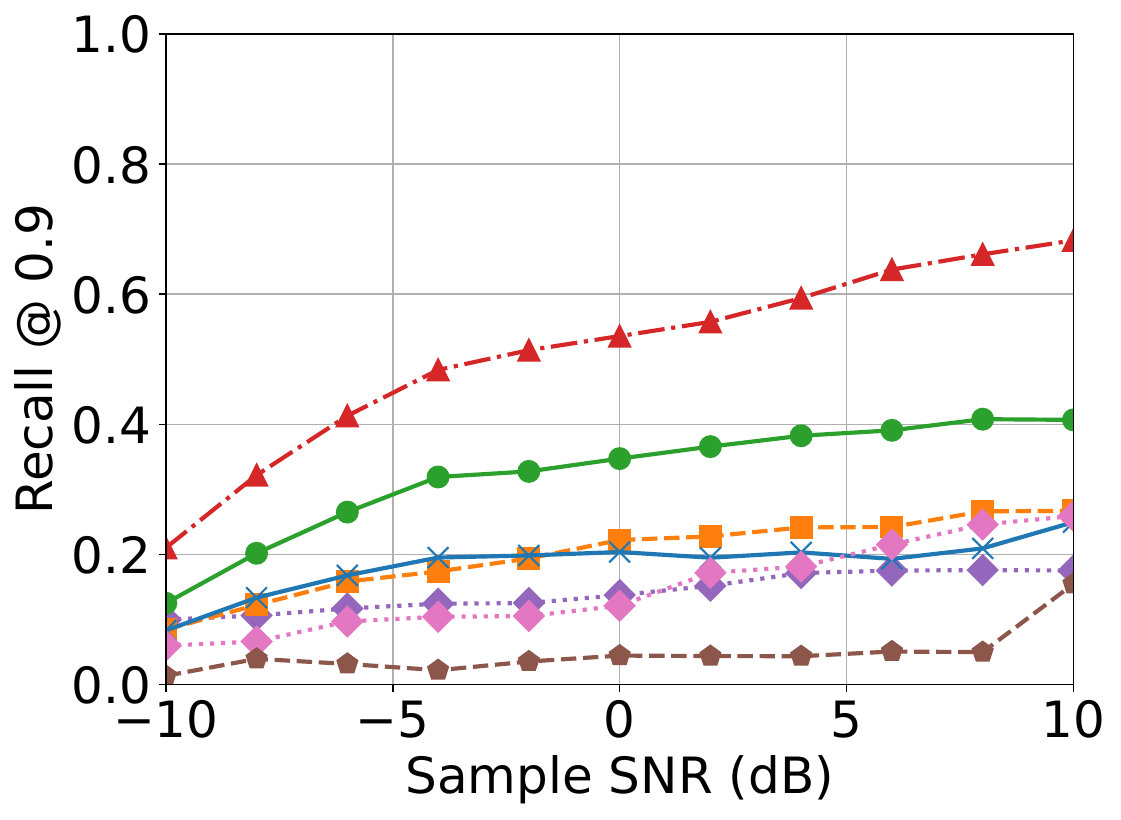}
  \caption{Recall (threshold=0.90)}
  \label{fig:OTA_Recall_0.9}
\end{subfigure}%
\vspace{-0.05in}
\caption{Indoor Over-the-air Dataset Spectrum segmentation performance under different SNR conditions.}
\label{fig:OTA_Performance}\vspace{-0.29in}
\end{figure*}
\vspace{-0.05in}
\subsection{Hyperparameter Analysis}
\label{subsec:hyper}

Hyperparameter selection is crucial for improving training and validation performance in $\mathbb{C}$MuSeNet. As we show next, a naive complex-valued conversion of real-valued neural networks is ineffective. Key parameters such as batch size, early stopping criteria, learning rate, and focal loss parameters ($\gamma, \alpha$) are evaluated. 
Final configurations are selected based on validation accuracy and training duration.

Batch size significantly influences validation accuracy and training time. Among configurations of $8$, $16$, $32$, and $64$, a batch size of $64$  achieves the highest validation accuracy of $99.43\%$ with a total training time of $5$ hours and $21$ minutes. In comparison, a batch size of $32$ achieves $99.0\%$ validation accuracy with a longer training time of $7$ hours and $38$ minutes. Smaller batch sizes further increase the training duration, with $16$ reaching $99.0\%$ in $11$ hours and $46$ minutes and $8$ achieving $98.7\%$ accuracy in $19$ hours and $24$ minutes. Larger batch sizes, such as $128$, are not tested due to GPU memory limitations. 

Early stopping is analyzed with patience values of $1$, $2$, $3$, $5$, and $10$. Early stopping halts training when the model fails to improve both validation accuracy and reduce validation loss within the specified patience, ensuring computational efficiency without compromising performance. A patience value of $3$ strikes a balance by preserving the maximum validation accuracy while minimizing the number of training epochs. Notably, lowering the patience to $2$ or $1$ results in premature stopping, preventing the model from reaching the peak accuracy achieved with a patience of $3$. Increasing the patience to $5$ or $10$ extends the training process by $8$ and $23$ additional epochs on average, respectively,  
without improving accuracy or reducing loss, resulting in unnecessary computational overhead. 
To explore the effect of complex-valued loss functions, we designed and tested complex-valued binary cross-entropy ($\mathbb{C}$BCE) alongside $\mathbb{C}$FL and their real-valued counterparts. $\mathbb{C}$BCE applies the BCE loss separately to the real and imaginary components as:
\begin{equation} \label{equation:CVBCE}
\begin{split}
L_{\mathbb{C}\text{BCE}} = &
-\frac{1}{2}\sum_{i \in \{x,y\}}\sum_{f=0}^{L-1} \Big[ o_{i,f}\log(p_{i,f}) \\
&+ (1 - o_{i,f})\log(1 - p_{i,f}) \Big],
\end{split}
 \end{equation} 
with the same parameter definitions in (\ref{equation:CVFL}). $\mathbb{C}$BCE achieves a validation accuracy of $98.4\%$, outperforming its real-valued counterpart $\mathbb{R}$BCE, which reaches $96.7\%$. On the other hand, $\mathbb{C}$FL demonstrates better accuracy, surpassing real-valued focal loss ($\mathbb{R}$FL), which achieves $97.1\%$ validation accuracy. Fine-tuning $\mathbb{C}$FL with $\gamma$ and $\alpha$ values from $1$ to $10$ results in accuracies between $99.11\%$ and $99.43\%$, with the configuration $\gamma = 1$ and $\alpha = 3$ selected for its balance of performance and efficiency. 
Utilizing IoU accuracy during training, the combination of $\mathbb{C}$IoU and $\mathbb{C}$FL achieves a validation accuracy of $99.43\%$, outperforming $\mathbb{R}$IoU, with an accuracy of $99.31\%$. This result underscores the effectiveness of incorporating $\mathbb{C}$FL and $\mathbb{C}$IoU within the training framework.

Further analysis involves removing the FFT step from the $\mathbb{C}$MuSeNet, and training the model directly on time-domain IQ signals. While the model can converge, it achieves a significantly lower validation accuracy of $89.13\%$ and requires $13$ hours and $42$ minutes to complete training (twice the time compared to $\mathbb{C}$MuSeNet with FFT), as discussed in Section~\ref{sec:eval}. This result highlights the importance of FFT preprocessing in achieving higher accuracy and improved training efficiency for spectrum segmentation. 

Based on the hyperparameter analysis, the final configuration of $\mathbb{C}$MuSeNet is selected as follows: a batch size of $64$, early stopping with patience of $3$ using accuracy measured with $\mathbb{C}$IoU as the threshold, and $\mathbb{C}$FL with $\gamma = 1$ and $\alpha = 3$, combined with FFT preprocessing. This selection balances training efficiency, convergence time, and segmentation performance, as demonstrated by the evaluations above. 
\vspace{-0.05in}
\subsection{Performance Comparison}
\vspace{-0.05in}
We evaluate the performance of $\mathbb{C}$MuSeNet and compare it with the following state-of-the-art deep learning and model-based solutions: (1) Real-valued ResNet22 with FFT preprocessing ($\mathbb{R}$ResNet22)~\cite{Li22ResNet22}, (2) real-valued ResNet18 with FFT preprocessing ($\mathbb{R}$ResNet18), which is the real-valued counterpart of $\mathbb{C}$MuSeNet architecture, (3) real-valued U-Net with FFT preprocessing ($\mathbb{R}$U-Net)~\cite{Subedi24}, (4) real-valued U-Net with power spectral density preprocessing ($\mathbb{R}$U-Net(PSD))~\cite{Subedi24}, (5) energy detection with localization algorithm based on double thresholding (LAD)~\cite{Vartiainen05LAD}. Finally, we design (6) a complex-valued counterpart of a U-Net architecture with FFT preprocessing ($\mathbb{C}$U-Net) to analyze the impacts of CVNN design on different architectures. Since PSD is a real-valued input, only FFT preprocessing is evaluated in $\mathbb{C}$U-Net. It is important to note that real-valued architecture can only process the absolute value of FFT results, while complex-valued architectures use complex-valued FFT results.
\subsubsection{Evaluation Metrics}
\vspace{-0.05in}
The performance of $\mathbb{C}$MuSeNet and the comparison models are evaluated using segmentation accuracy at an IoU threshold of $0.5$, IoU scores, recall, as well as epoch time and total training time. For fairness, for both CVNN and RVNN models, IoU scores, and thresholds are calculated using a real-valued IoU metric: $IoU = \{min(\hat{f^e}, f^e)-max(\hat{f^b}, f^b)\}/\{max(\hat{f^e}, f^e)-min(\hat{f^b}, f^b)\}$~\cite{Subedi24}. For CVNN models, the real-valued IoU is applied to the absolute values of the complex-valued outputs. This ensures consistent evaluation across both CVNN and RVNN models. Here, it is important to note that while evaluations utilize real-valued IoU for a fair comparison, the complex-valued loss functions leverage the \textit{complex-valued} IoU, which we introduce in (\ref{equation:CVIoU}) during training. Recall is assessed at IoU thresholds of $0.5$ and $0.9$, capturing general and precise segmentation performance. Training efficiency is evaluated by measuring the time taken for each epoch (epoch time) and the total number of epochs required to complete training. 
All metrics are computed using an input size of $16,384$.

The performance of $\mathbb{C}$MuSeNet and the comparison models are summarized in Table~\ref{table:Evaluation_Result} across the three datasets: Synthetic, indoor OTA and BIG-RED. 

\begin{table}[t!]
    \centering
    \caption{Average accuracy, IoU score, and recall over SNR values for synthetic, indoor OTA, and BIG-RED datasets.} \vspace{-0.05in}
    \begin{adjustbox}{width=0.49\textwidth}
        \begin{tabular}{|l|c|c|c|c|}
            \hline
            \multicolumn{5}{|c|}{Synthetic} \\ \hline
                                    & Accuracy (\%) & IoU Score & Recall @0.5 & Recall @0.9 \\ \hline
            $\mathbb{C}$MuSeNet     & \textbf{99.40}     & \textbf{0.92}  & 0.97             & \textbf{0.79}    \\ \hline
            $\mathbb{R}$U-Net(PSD)~\cite{Subedi24} & 98.97              & \textbf{0.92}  & \textbf{0.99}    & 0.77 \\ \hline
            $\mathbb{C}$U-Net       & 96.54              &  0.68          & 0.75             & 0.53 \\ \hline
            $\mathbb{R}$U-Net~\cite{Subedi24}       & 96.05              &  0.65          & 0.71             & 0.51 \\ \hline
            $\mathbb{R}$ResNet22~\cite{Li22ResNet22}    & 93.48              &  0.73          & 0.86             & 0.31 \\ \hline
            $\mathbb{R}$ResNet18    & 92.46              &  0.72          & 0.84             & 0.22 \\ \hline
            LAD~\cite{Vartiainen05LAD}                     & 80.12              &  0.45          & 0.52             & 0.04\\ \hline
            \multicolumn{5}{|c|}{Indoor Over-the-air} \\ \hline
                                    & Accuracy (\%) & IoU Score & Recall @0.5 & Recall @0.9 \\ \hline
            $\mathbb{C}$MuSeNet     & \textbf{98.98}     & \textbf{0.81}  & \textbf{0.90}    & \textbf{0.53}    \\ \hline
            $\mathbb{R}$U-Net(PSD)~\cite{Subedi24} & 97.42              & 0.70           & 0.62             & 0.19              \\ \hline
            $\mathbb{C}$U-Net       & 96.46              & 0.63           & 0.61             & 0.33  \\ \hline
            $\mathbb{R}$U-Net~\cite{Subedi24}       & 95.80              & 0.56           & 0.53             & 0.21  \\ \hline
            $\mathbb{R}$ResNet22~\cite{Li22ResNet22}    & 95.88              & 0.58           & 0.67             & 0.15  \\ \hline
            $\mathbb{R}$ResNet18    & 95.18              & 0.58           & 0.63             & 0.14  \\ \hline
            LAD~\cite{Vartiainen05LAD}                     & 89.96              & 0.48           & 0.54             & 0.05  \\ \hline
            \multicolumn{5}{|c|}{BIG-RED} \\ \hline
                                    & Accuracy (\%) & IoU Score & Recall @0.5 & Recall @0.9 \\ \hline
            $\mathbb{C}$MuSeNet     & \textbf{99.90}     & \textbf{0.95}  & \textbf{0.97}    & \textbf{0.75}     \\ \hline
            $\mathbb{R}$U-Net(PSD)~\cite{Subedi24} & 96.77              & 0.75           & 0.78             & 0.22  \\ \hline
            $\mathbb{C}$U-Net       & 95.42              & 0.61           & 0.64             & 0.19  \\ \hline
            $\mathbb{R}$U-Net~\cite{Subedi24}       & 96.62              & 0.66           & 0.61             & 0.20  \\ \hline
            $\mathbb{R}$ResNet22~\cite{Li22ResNet22}    & 91.60              & 0.60           & 0.46             & 0.16  \\ \hline
            $\mathbb{R}$ResNet18    & 90.70              & 0.54           & 0.45             & 0.12  \\ \hline
            LAD~\cite{Vartiainen05LAD}                     & 85.40              & 0.52           & 0.48             & 0.02  \\ \hline
        \end{tabular} \vspace{-0.05in}
    \end{adjustbox}
    \label{table:Evaluation_Result}\vspace{-0.25in}
\end{table}

\subsubsection{Synthetic Dataset Evaluations}
The segmentation performance across different sample SNR values is shown in Figs.~\ref{fig:Syn_Performance}. $\mathbb{C}$MuSeNet achieves an average accuracy of $99.40\%$, and an average IoU score of $0.92$. The performance remains consistent down to SNR of $-16$ dB, with IoU decreasing slightly to $0.88$ at SNR of $-20$ dB as shown in Fig.~\ref{fig:Syn_IoU}. The average recall at an IoU threshold of $0.5$ is $0.97$, while at a stricter threshold of $0.9$, the recall is $0.79$. 
Compared to the corresponding RVNN model $\mathbb{R}$ResNet18, $\mathbb{C}$MuSeNet has a $6.94$ percentage point (ppt) higher average accuracy and a $0.20$ higher IoU score. Recall at $0.5$ threshold is $0.13$ higher, and at a stricter $0.9$ threshold, $\mathbb{C}$MuSeNet outperforms $\mathbb{R}$ResNet18 with a substantial $0.57$ difference. These results show that the complex-valued architecture and the proposed loss function improve overall segmentation accuracy compared to RVNN, resulting in a more accurate assessment of the available spectrum in dynamic spectrum access scenarios.

Compared to the state-of-the-art $\mathbb{R}$U-Net (PSD)~\cite{Subedi24} and $\mathbb{R}$ResNet22~\cite{Li22ResNet22}, $\mathbb{C}$MuSeNet improves average accuracy by $0.5$ ppt and $5.92$ ppt, respectively. This difference becomes more pronounced in the low-SNR regime, where $\mathbb{C}$MuSeNet maintains $98.7\%$ average accuracy, while $\mathbb{R}$U-Net and $\mathbb{R}$ResNet22 decreases to $95.9\%$ and $82.7\%$, resulting in a gap of $2.8$ ppt and $16$ ppt, respectively. As shown in Fig.~\ref{fig:Syn_Accuracy}, $\mathbb{C}$MuSeNet exhibits only a slight degradation at SNR of $-20$ dB, whereas both state-of-the-art suffers a significant drop in performance.
The performance gap is further evident in the recall at the IoU threshold of $0.9$. While the average recall difference is $0.02$, $\mathbb{R}$U-Net (PSD) declines steeply below an SNR of $-12$ dB, reaching as low as $0.55$ and  $\mathbb{R}$ResNet22 reaching significantly lower at $0.19$ at an SNR of $-20$ dB, as illustrated in Fig.~\ref{fig:Syn_Recall_0.9}. Consequently, $\mathbb{C}$MuSeNet can continue to provide high-quality segmentation for weak signals.

Although $\mathbb{C}$MuSeNet demonstrates superior performance over RVNN models in the synthetic dataset, the differences are less pronounced due to the limited complexity of the AWGN channel. The synthetic dataset lacks the intricate channel characteristics and imperfections found in real-world environments. To this end, we next evaluate the indoor OTA dataset, where such complexities are present. 
\begin{figure*}[t!]
\centering
\begin{subfigure}{0.62\columnwidth}
  \centering
  \includegraphics[width=\textwidth]{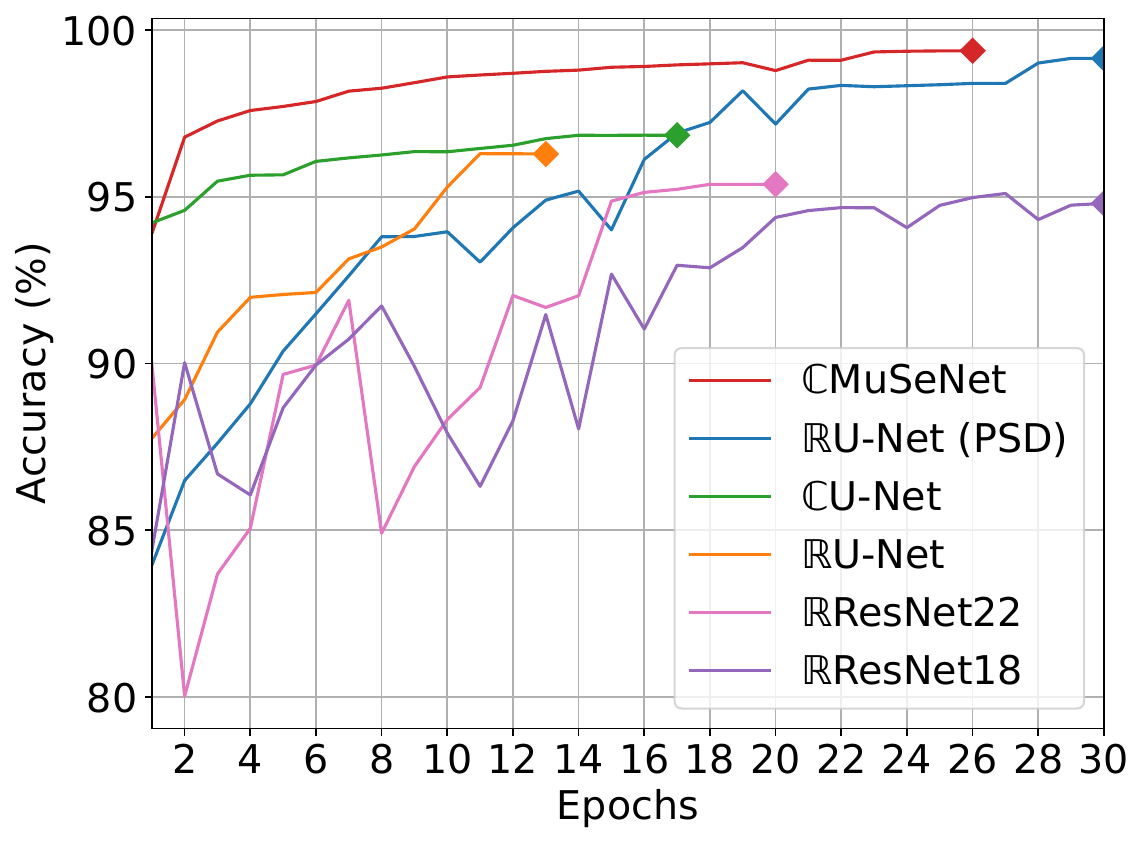}
  \caption{Training efficiency comparison}
  \label{fig:accuracy_convergence}
\end{subfigure}%
\begin{subfigure}[b]{0.62\columnwidth}
  \centering
  \includegraphics[width=\textwidth]{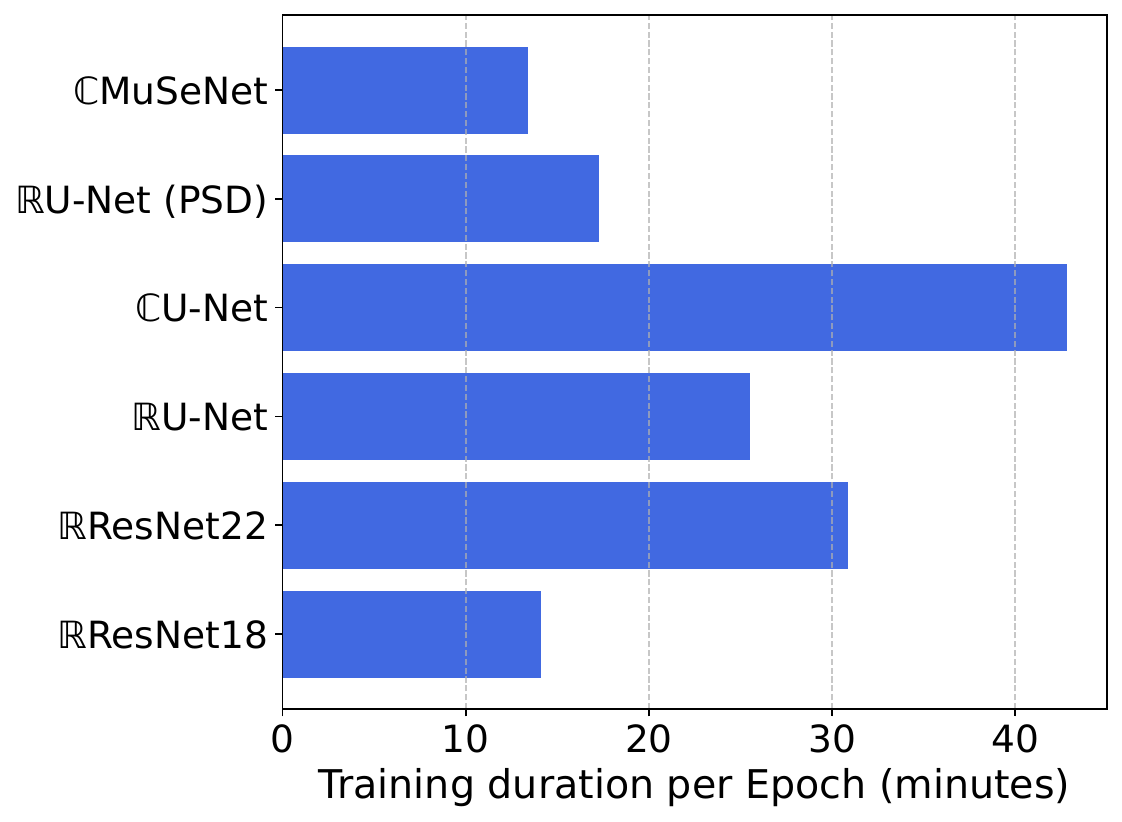}
  \caption{Average training duration per epoch}
  \label{fig:epoch_duration}
\end{subfigure}%
\begin{subfigure}[b]{0.62\columnwidth}
  \centering
  \includegraphics[width=\textwidth]{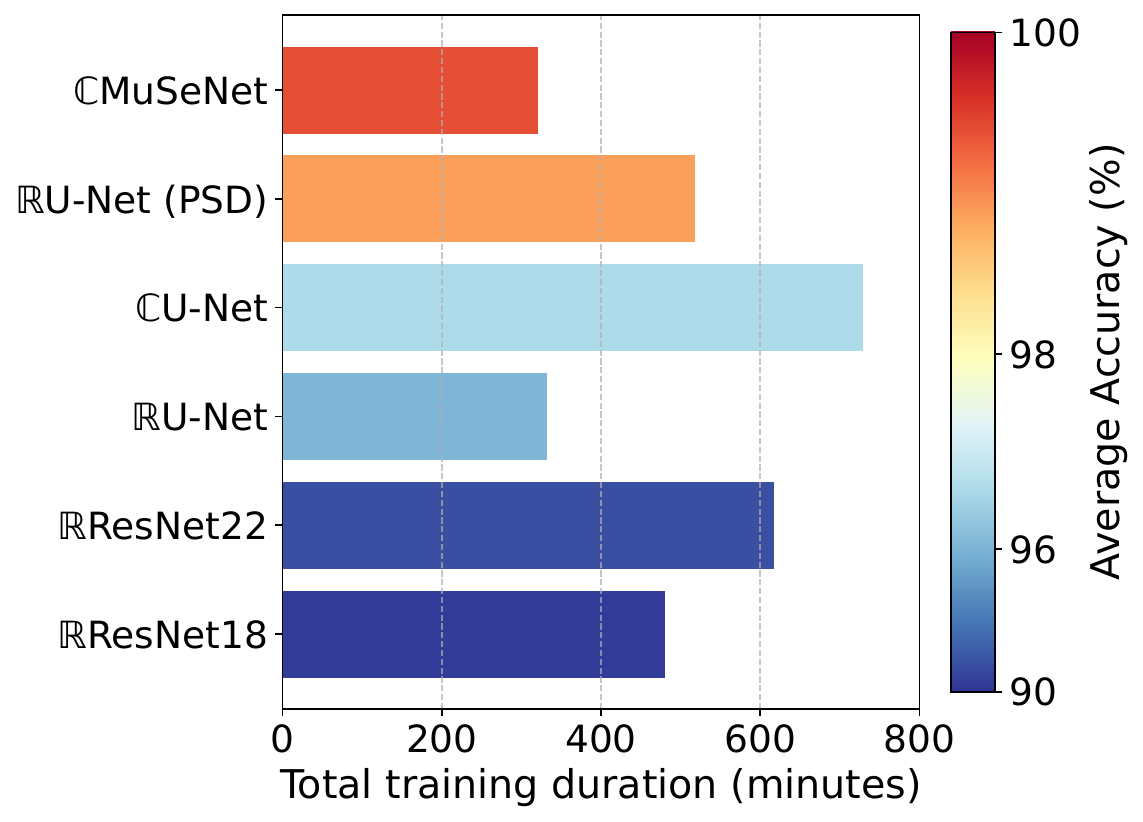}
  \caption{Total training time with accuracy}
  \label{fig:total_duration}
\end{subfigure}%
\vspace{-0.05in}
\caption{Training efficiency comparison}
\label{fig:Training_Efficiency}\vspace{-0.3in}
\end{figure*}
\subsubsection{Indoor OTA Dataset Evaluations}
$\mathbb{C}$MuSeNet's performance across varying SNR levels is illustrated in Figs.~\ref{fig:OTA_Performance}. With an average accuracy of $98.98\%$, $\mathbb{C}$MuSeNet surpasses $\mathbb{R}$ResNet18 by $3.8$ ppt, $\mathbb{R}$ResNet22 by $3.1$ ppt, $\mathbb{R}$U-Net by $3.18$ ppt, and $\mathbb{R}$U-Net (PSD) by $1.42$ ppt. Notably, $\mathbb{C}$MuSeNet maintains over $97.54\%$ accuracy across low-SNR scenarios such as $-10$ dB, where other models experience substantial drops as illustrated in Fig.~\ref{fig:OTA_Accuracy}. For example, $\mathbb{R}$U-Net (PSD), $\mathbb{R}$U-Net, and $\mathbb{R}$ResNet22 accuracy drops down to $95.22\%$, $94.04\%$, and $92.51\%$, respectively.  
For IoU scores, as shown in Fig.~\ref{fig:OTA_IoU}, $\mathbb{C}$MuSeNet shows consistent improvement across SNR levels, reaching $0.67$ at $-10$ dB, compared to $0.43$ for $\mathbb{R}$U-Net, $0.58$ for $\mathbb{R}$U-Net (PSD), and $0.45$ for $\mathbb{R}$ResNet22.
Recall at an IoU threshold of 0.5 remains above $0.78$ for $\mathbb{C}$MuSeNet across all SNR values as illustrated in Fig.~\ref{fig:OTA_Recall_0.5}. At an SNR of $-10$ dB, $\mathbb{C}$MuSeNet achieves $0.28$ higher score than $\mathbb{R}$ResNet22, $0.47$ higher than $\mathbb{R}$U-Net, and $0.1$ higher than $\mathbb{R}$U-Net (PSD). For the stricter IoU threshold of $0.9$, the difference between $\mathbb{C}$MuSeNet and the comparison model is significant in the low-SNR regime, as shown in Fig.~\ref{fig:OTA_Recall_0.9}. $\mathbb{C}$MuSeNet achieves $0.25$ at $-10$ dB, compared to $0.09$ for $\mathbb{R}$U-Net (PSD), $0.06$ for $\mathbb{R}$ResNet22, and $0.7$ of the $\mathbb{R}$U-Net. The comparative results demonstrate that $\mathbb{C}$MuSeNet consistently outperforms state-of-the-art real-valued models such as $\mathbb{R}$ResNet22 and $\mathbb{R}$U-Net (PSD) across all SNR levels, with particularly significant improvements observed at low-SNRs. 

It is important to note that while $\mathbb{C}$U-Net outperforms its real-valued counterpart, this architecture cannot reach the accuracy levels of $\mathbb{C}$MuSeNet, illustrating the importance of architecture choice in CVNNs.
These findings further emphasize the robustness of $\mathbb{C}$MuSeNet in handling diverse channel effects and transmission conditions. Building on this, we now evaluate the framework on the BIG-RED, representing extreme and highly challenging real-world scenarios. 
\subsubsection{BIG-RED Evaluations}
The BIG-RED is a diverse spectrum dataset, introducing dynamic channel conditions, varying signal strengths, and environmental noise. Due to the dataset's extensive coverage and realistic conditions, it is evaluated using averaged metrics rather than SNR-specific comparisons. 
$\mathbb{C}$MuSeNet achieves the highest average accuracy of $99.90\%$, significantly outperforming $\mathbb{R}$ResNet18, which achieves $90.7\%$, $\mathbb{R}$U-Net at $96.62\%$, and $\mathbb{R}$U-Net (PSD) at $96.77\%$. These results highlight the effectiveness of the CVNN-based architecture in managing the raw and dynamic characteristics of wild RF data. Note that the subset of the BIG-RED used in this evaluation benefits from a high-gain reception setup in NEXTT. This ensures higher signal fidelity in the received samples and provides a strong foundation for model evaluation, supporting the observed high accuracy despite the dataset's complexity. 
For average IoU scores, $\mathbb{C}$MuSeNet achieves $0.95$, a substantial improvement over $\mathbb{R}$ResNet18 at $0.54$, $\mathbb{R}$U-Net at $0.56$ and $\mathbb{R}$U-Net (PSD) at $0.70$. This considerable advantage in IoU scores underscores the model's capability to accurately identify and segment signal boundaries, even in highly unstructured and complex samples.
At an IoU threshold of $0.5$, $\mathbb{C}$MuSeNet achieves an average recall of $0.97$, far surpassing $\mathbb{R}$ResNet18 at $0.45$, $\mathbb{R}$U-Net at $0.78$. At the stricter threshold of $0.9$, $\mathbb{C}$MuSenet achieves a recall of $0.75$, compared to $0.12$ for $\mathbb{R}$ResNet18, $0.20$ for $\mathbb{R}$U-Net, and $0.22$ for $\mathbb{R}$U-Net (PSD).
These results illustrate $\mathbb{C}$MuSeNet's consistent segmentation performance across complex and diverse signal environments and its agility in real-world spectrum scenarios. We also compare $\mathbb{C}$MuSeNet with a commonly used energy detection algorithm, LAD, demonstrating that $\mathbb{C}$MuSeNet significantly 
outperforms LAD by up to $19.28$ ppt in average accuracy, $0.47$ in average IoU score, and $0.75$ in average recall at a strict $0.9$ IoU threshold across the evaluated datasets. Building on these findings, we now analyze the training efficiency of the framework, focusing on training time and computational performance. 
\vspace{-0.07in}
\subsection{Training Efficiency}
\vspace{-0.05in}
The training efficiency of CVNNs compared to RVNNs remains a debated topic in the research community~\cite{Wu23CVNNMorebutSlow, Xu22AnalysisCVNN_RF}. To address this issue, we evaluate the training efficiency of $\mathbb{C}$MuSeNet, focusing on three critical aspects. First, we examine the number of epochs required for $\mathbb{C}$MuSeNet to reach the maximum validation accuracy achieved by RVNNs. Second, we assess the average training duration per epoch to quantify the computational cost of CVNN operations compared to RVNNs. Finally, we measure the total training time required to complete the training process, providing an overall comparison of computational efficiency. 
\subsubsection{Epochs to Maximum Validation Accuracy}
The convergence trends for all evaluation models are illustrated in Fig.~\ref{fig:accuracy_convergence}, showcasing the number of epochs required to achieve maximum validation accuracies. Among the models, $\mathbb{C}$MuSeNet has the fastest convergence, surpassing the maximum validation accuracy of $\mathbb{R}$ResNet18 $(95.10\%)$ and $\mathbb{R}$ResNet22 $(95.37\%)$ within just $2$ epochs. In contrast, $\mathbb{R}$ResNet18 requires $27$ epochs, and $\mathbb{R}$ResNet22 requires $18$ epochs to reach their peak accuracy. For the U-Net-based models, $\mathbb{C}$U-Net achieves the maximum validation accuracy of $\mathbb{R}$U-Net in $7$ epochs, compared to the $13$ epochs required by $\mathbb{R}$U-Net. Similarly, $\mathbb{R}$U-Net (PSD) reaches its maximum accuracy in $30$ epochs, but $\mathbb{C}$U-Net still converges faster. The results clearly illustrate the advantages of $\mathbb{C}$MuSeNet and $\mathbb{C}$U-Net in learning efficiency compared to their real-valued counterparts.
The faster convergence of CVNN-based models like $\mathbb{C}$MuSeNet and $\mathbb{C}$U-Net can be attributed to their ability to process complex-valued data directly. The inherent design of CVNNs allows for more efficient feature extraction and learning, particularly when processing complex-valued Fourier-transformed IQ signals.
Having established the faster convergence properties of CVNN-based architectures, we now turn to the practical training time by analyzing the average training duration per epoch. 
\vspace{-0.02in}
\subsubsection{Average Training Duration per Epoch}
The average training duration per epoch provides insight into the time efficiency of CVNN and RVNN models during the training process. As shown in Fig.~\ref{fig:epoch_duration}, $\mathbb{C}$MuSeNet requires $802$ seconds per epoch, compared to $846$ seconds for $\mathbb{R}$ResNet18 ($5.5\%$ faster) and $1,850$ seconds for  $\mathbb{R}$ResNet22 ($130.7\%$ faster). This challenges the assumption that CVNNs are inherently slower due to their complex-valued operations and aligns with findings in~\cite{Xu22AnalysisCVNN_RF} that demonstrate competitive training times for CVNNs in ResNet-like architectures.
For the U-Net-based models, $\mathbb{C}$U-Net takes $2,570$ seconds per epoch, compared to $1,530$ seconds for $\mathbb{R}$U-Net ($68\%$ slower) and $1,032$ seconds for $\mathbb{R}$U-Net (PSD) ($149\%$ slower). 
These findings emphasize the importance of architectural design in leveraging CVNN benefits effectively. While CVNNs excel in preserving IQ signal characteristics, the choice of architecture and its suitability for CVNN adaptation play a critical role in balancing training time and performance. 
To fully evaluate training efficiency, we now analyze the total training duration for each model.

\subsubsection{Total Training Duration}
The total training duration provides a comprehensive measure of the time required to fully train each model, considering both the average per-epoch duration and the total number of epochs needed for convergence. As shown in Fig.~\ref{fig:total_duration}, $\mathbb{C}$MuSeNet completes training in $5$ hours and $21$ minutes, significantly faster than the $8$ hours required by $\mathbb{R}$ResNet18 ($33.1\%$ faster) and $10$ hours $16$ minutes required by $\mathbb{R}$ResNet22 ($92.2\%$ faster).
For the U-Net based models, $\mathbb{C}$U-Net takes $12$ hours and $9$ minutes, compared to $8$ hours and $36$ minutes for $\mathbb{R}$U-Net (PSD). While $\mathbb{C}$U-Net demonstrates improved segmentation performance over its RVNN counterparts, the increased training duration highlights the need for careful architectural design when applying CVNNs. The results underscore that while CVNNs consistently improve performance across all architectures, the efficiency and practicality of their application depend on the balance between performance gains and computational demands. 
$\mathbb{C}$MuSeNet demonstrates a clear advantage by achieving high performance and faster training times compared to RVNN-based models. It consistently outperforms the state-of-the-art in both synthetic and real-world datasets. Importantly, the results show that CVNNs enhance segmentation accuracy across all architectures, with frameworks like $\mathbb{C}$MuSeNet achieving these gains with \textit{lower} computational overhead. This balance of efficiency and accuracy positions $\mathbb{C}$MuSeNet as a practical and effective solution for spectrum segmentation, particularly in challenging conditions such as those represented by BIG-RED.
\section{Conclusion}
\label{sec:conclusion}
This work presents $\mathbb{C}$MuSeNet, a novel complex-valued residual network architecture for multi-signal segmentation in wideband spectrum sensing. Leveraging the inherent advantages of CVNNs, $\mathbb{C}$MuSeNet processes IQ samples directly, preserving critical phase and amplitude information that traditional real-valued models fail to capture fully. By integrating advanced $\mathbb{C}$FL function and $\mathbb{C}$IoU, the framework outperforms the state-of-the-art in terms of both accuracy and training time. 
The evaluations demonstrate that $\mathbb{C}$MuSeNet consistently outperforms state-of-the-art real-valued models across a wide range of datasets, including synthetic, indoor, and city-wide outdoor datasets. Notably, $\mathbb{C}$MuSeNet achieves an average accuracy of $98.98 \% - 99.90\%$, highlighting its adaptability to diverse conditions. $\mathbb{C}$MuSeNet maintains high IoU and recall across varying scenarios, particularly in low-SNR and real-world environments. This performance underscores the robustness and practicality of CVNNs for spectrum segmentation. Additionally, $\mathbb{C}$MuSeNet training is up to $92.2\%$ faster, making it suitable for real-world applications.

To the best of our knowledge, this is the first \textit{complex-valued} neural network-based multi-signal spectrum segmentation framework designed for cognitive radio systems. $\mathbb{C}$MuSeNet lays the groundwork for future spectrum sensing and management research, offering a new paradigm for processing complex-valued data in dynamic wireless environments. Future work will focus on extending its scalability and enhancing real-time processing. Additionally, we will explore deployment in distributed cognitive radio networks, further advancing the capabilities of next-generation spectrum management systems.
\vspace{-0.1in}

\bibliographystyle{IEEEtranS}

\end{document}